\documentclass{article}

\usepackage{amssymb,amsthm,amsmath} 
\usepackage{enumerate}
\usepackage{graphicx}
\usepackage{enumitem}
\usepackage{url}

\usepackage{array}
\usepackage{booktabs}

\usepackage[a-3b]{pdfx}
\usepackage[pdfa]{hyperref}
\usepackage{lipsum}
\usepackage{geometry}
\usepackage{todonotes} 

\theoremstyle{definition}

\newtheorem{algoritmi}{Algorithm}[section]
\newtheorem{theorem}{Theorem}[section]
\newtheorem{assumption}{Assumption}[section]
\newtheorem{remark}{Remark}[section]

\theoremstyle{plain}

\newcommand{\R}{\mathbb{R}}
\newcommand{\ba}{{\boldsymbol a}} 
\newcommand{\x}{{\boldsymbol x}} 
\newcommand{\y}{{\boldsymbol y}} 
\newcommand{\z}{{\boldsymbol z}} 

\begin{document}

\title{Clust-Splitter $-$ an Efficient Nonsmooth Optimization-Based Algorithm for Clustering Large Datasets}







\author{
  Jenni Lampainen\textsuperscript{1*},\;
  Kaisa Joki\textsuperscript{1},\;
  Napsu Karmitsa\textsuperscript{2},\;
  Marko M. Mäkelä\textsuperscript{1}\\[1.5ex]
  \textsuperscript{1}\small Department of Mathematics and Statistics, University of Turku, FI-20014 Turku, Finland\\[-0.7ex]
  \textsuperscript{2}\small Department of Computing, University of Turku, FI-20014 Turku, Finland
}


\date{}

\maketitle

\vspace{-2em}

\begin{center}
{$^*$Corresponding author. E-mail:} \href{mailto:jmlamp@utu.fi}{jmlamp@utu.fi};

{Contributing authors:}
\href{mailto:kaisa.joki@utu.fi}{kaisa.joki@utu.fi}; \href{mailto:napsu@karmitsa.fi}{napsu@karmitsa.fi}; \href{mailto:makela@utu.fi}{makela@utu.fi};
\end{center}

\vspace{1.5em}

\begin{abstract}
\noindent Clustering is a fundamental task in data mining and machine learning, particularly for analyzing large-scale data. In this paper, we introduce \textsc{Clust-Splitter}, an efficient algorithm based on nonsmooth optimization, designed to solve the minimum sum-of-squares clustering problem in very large datasets. The clustering task is approached through a sequence of three nonsmooth optimization problems: two auxiliary problems used to generate suitable starting points, followed by a main clustering formulation. To solve these problems effectively, the limited memory bundle method is combined with an incremental approach to develop the \textsc{Clust-Splitter} algorithm. We evaluate \textsc{Clust-Splitter} on real-world datasets characterized by both a large number of attributes and a large number of data points and compare its performance with several state-of-the-art large-scale clustering algorithms. Experimental results demonstrate the efficiency of the proposed method for clustering very large datasets, as well as the high quality of its solutions, which are on par with those of the best existing methods.\\

\noindent \textbf{Keywords:} Clustering, Nonsmooth optimization, Nonconvex optimization, Limited memory bundle method, Incremental algorithm, Large-scale data

\end{abstract}

\medskip

\begingroup
\renewcommand{\thefootnote}{}  
\footnotetext{\textbf{Note:} This work has been submitted to the IEEE for possible publication. Copyright may be transferred without notice, after which this version may no longer be accessible.}
\addtocounter{footnote}{-1}    
\endgroup

\section{Introduction}

Clustering is a fundamental task in data mining and machine learning that organizes data points into clusters based on their similarity. It plays a vital role in applications such as bioinformatics \cite{bioinf1, bioinf2}, cybersecurity \cite{cybersec1, cybersec2}, and image processing \cite{9151332}. This paper develops a method for \emph{the hard clustering problem}, where each data point is assigned to exactly one cluster.

A key aspect of cluster analysis is the choice of a similarity measure, which can be defined using different norms. In our work, we adopt the widely accepted squared Euclidean norm. Therefore, the clustering problem we address is referred to as \emph{the minimum sum-of-squares clustering (MSSC) problem}, where the goal is to partition data points into clusters minimizing the sum of squared distances to the cluster centers.

The MSSC problem can be expressed as a global optimization problem, and various optimization techniques have been applied to develop clustering algorithms for solving it. These techniques include:
\begin{itemize}
    \item nonsmooth optimization methods \cite{BagAliSUl2023, BagKarTah:2025, BagMoh:2015, BagUgo:2005, BagYea:2006, KarBagTah:2017, KarBagTah:2018};
    \item methods based on the difference of convex (DC) representation of the clustering function \cite{BagKarTah:2025, BagTahUgo:2016, KarBagTah:2017, KhaAstDAlGau:2016, LeTBelPha:2007, LeTLeHPha:2014, LeTPha:2009} and global optimization \cite{Xia:2009};
    \item the hyperbolic smoothing technique \cite{BagOrdOztXav:2015, Xav:2010, XavXav:2011, XavXav2020}; 
    \item the variable neighborhood search algorithm \cite{HanMla:2001}; and
    \item metaheuristics such as simulated annealing \cite{SeiBagBorPic2019, SelAlS:1991}, tabu search \cite{Al-Sultan1995, tabu1}, evolutionary algorithms \cite{XuKliHer:2014}, particle swarm optimization \cite{PSO1, PSO2}, and genetic algorithms \cite{GriVid2019, ManSch2021, rahman2014}.
\end{itemize}
In addition, heuristic methods such as the $k$-means algorithm and its variations are commonly used to address the MSSC problem (see, e.g., \cite{Bag:2008, Jai:2010, VolTolWeb:2013}).

Recent advancements in computer hardware allow for the storage of large datasets, including as many as millions of data points and attributes, in random-access memory (RAM). This capability makes it possible to process massive datasets at once, enabling accurate clustering results. However, many existing clustering algorithms are not suitable for such large datasets, as they either produce suboptimal outcomes, such as local minima, or require excessive computational resources. Therefore, new clustering algorithms capable of generating accurate results within a reasonable time for very large datasets are needed.

Using nonsmooth optimization (NSO) \cite{BagKarMak:2014} in clustering provides a solution to this problem. In traditional clustering methods, the number of variables often increases with the number of data points. In contrast, NSO employs a fixed number of variables determined by the number of attributes and clusters. This significantly reduces the dimensionality of the optimization problem, decreasing computational complexity and making NSO particularly well-suited for large-scale clustering tasks. However, methods based on global optimization techniques are often computationally too intensive. On the other hand, methods relying on efficient local optimization techniques require a good starting point in order to find a global or even a good local solution. A common strategy to address this is to use an incremental approach, fundamentally based on the starting point selection procedure introduced in \cite{OrdBag:2015}. This approach, or a slight variation of it, is used, for example, in \cite{BagAliSUl2023, BagMoh:2015, BagOrdOztXav:2015, BagTahUgo:2016, KarBagTah:2017, KarBagTah:2018, KhaAstDAlGau:2016, XavXav2020}.

In this paper, we introduce a novel incremental clustering method, \textsc{Clust-Splitter}, for solving large-scale MSSC problems. \textsc{Clust-Splitter} is based on the NSO approach and incorporates a new splitting strategy to generate starting points. The core of this method lies in data splitting: at each iteration, once a solution is obtained for the current number of clusters, the cluster with the highest dissimilarity is split to generate effective starting points for the main clustering task with one additional cluster. This unique strategy of splitting clusters based on their similarity clearly distinguishes \textsc{Clust-Splitter} from the other incremental clustering methods mentioned above. In particular, the proposed approach is computationally more efficient and the resulting optimization problems are significantly easier to solve. To solve the clustering problem, we employ the limited memory bundle method (LMBM) \cite{Haa:2004, HaaMieMak:2004, HaaMieMak:2007}, which is considered an efficient method for solving large-scale NSO problems. Moreover, the LMBM has been successfully applied in various machine learning applications (see, e.g., \cite{KarBagTah:2018, KarTahBag:2022, KarTahJok:2023}).

The main aims of this paper are:
\begin{itemize}
    \item[(i)] To develop a novel efficient incremental algorithm for solving MSSC problems.
    \item[(ii)] To design an algorithm achieving high accuracy and efficiency in solving clustering problems for datasets with hundreds of thousands of data points or hundreds of features.
    \item[(iii)] To conduct a numerical analysis of the proposed \textsc{Clust-Splitter} method and compare its performance with \textsc{LMBM-Clust} \cite{KarBagTah:2018}, \textsc{DC-Clust} \cite{BagTahUgo:2016}, \textsc{Big-Clust} \cite{BigClust:2025}, \textsc{Big-Means} \cite{MusMlaJarMus2023}, and \textsc{MiniBatchKMeans} \cite{scikit} on very large datasets.
\end{itemize}

The structure of this paper is as follows. Section \ref{kappale2} outlines the notations and fundamental concepts from nonsmooth analysis. In Section \ref{kappale3}, formulations of nonsmooth clustering problems are presented. A detailed description of the proposed new incremental clustering algorithm is provided in Section \ref{kappale4}. Section \ref{kappale5} discusses the results of numerical experiments. Finally, some conclusions of the study are presented in Section \ref{kappale6}. The Appendix includes detailed numerical results presented in tables and figures.

\section{Notations and background} \label{kappale2}

A function $f : \R^n \to \R $ is \emph{nonsmooth} if there exists at least one point $\x \in \R^n$ where $f$ is not continuously differentiable. We use $\|\cdot\|$ to denote the Euclidean norm in $\R^n$. Throughout, bold symbols represent vectors.

A function $f: \R^n \to \R$ is said to be \emph{convex} if
$$f(\lambda \x + (1-\lambda) \y) \leq \lambda f(\x) + (1-\lambda) f(\x) \quad \text{ for all } \; \x, \y \in \R^n \text{ and } \lambda \in [0,1].$$
In addition, a function $f: \R^n \to \R$ is called \emph{locally Lipschitz continuous} (LLC) on $\mathbb{R}^n$ if for any bounded subset $X \subset \R^n$ there exists a constant $L > 0$ such that 
$$|f(\x) - f(\y)| \leq L \|\x - \y \| \quad \text{ for all } \; \x, \y \in X.$$
Note that a convex function is always LLC \cite{BagKarMak:2014}.

The \emph{Clarke subdifferential} of an LLC function $f: \R^n \to \R$ at any point $\x \in \R^n$ is defined as \cite{Cla:1983}
$$
\partial f(\x) = \operatorname{conv} \left\{ \lim_{i \to \infty} \nabla f(\x_i) \;|\; \x_i \to \x \; \text{ and } \; \nabla f(\x_i) \text{ exists} \right\},
$$
where ``$\operatorname{conv}$" refers to the convex hull of a set. A vector $\boldsymbol{\xi} \in \partial f(\x)$ is called a \emph{subgradient} whereas a point $\x^* \in \R^n$ is termed \emph{stationary} if $\mathbf{0} \in \partial f(\x^*)$. Stationarity is a necessary condition for local optimality. If the function $f$ is convex, stationarity becomes a sufficient condition for global optimality \cite{BagKarMak:2014, Cla:1983}.

\section{Nonsmooth formulations of clustering problems} \label{kappale3}

In this section, we introduce the NSO formulations for both the main clustering problem and the associated auxiliary clustering problems, which are used to generate suitable starting points to solve the main problem. We also review the definitions of clustering and similarity measure. The notations used in this paper are summarized in Table \ref{table_clustering_notations}.

\begin{table}[ht!]
    \centering
    \caption{Notations for clustering.} \vspace{0.25cm}
    \label{table_clustering_notations}
    \begin{tabular}{@{}ll@{}}
        \toprule
        \midrule
        \( m \)        & number of data points (i.e., records) \\
        \( n \)        & number of features \\
        \( k \)        & number of clusters \\
        \( k_{max} \)        & maximum number of clusters \\
        \( \ba^i \in \R^n \)        & data point, $i=1, \ldots, m$ \\
        \( A = \{\ba^1, \ldots, \ba^m \} \)        & dataset \\
        \( A^j \subseteq A \)        & cluster, $j=1,\ldots,k$ \\
        \( \x^j \in \R^n \)        & cluster center, $j = 1, \ldots, k$ \\
        \( f_k( \x) \)        & $k$-clustering function value at $\x=( \x^1, \ldots, \x^k) \in \R^{nk}$ \\
        \( \tilde{f}( \z) \)        & starting point auxiliary function value at $\z \in \R^n$ \\
        \( \hat{f}( \y) \)        & 2-clustering auxiliary function value at $\y = ( \y^1, \y^2) \in \R^{2n}$ \\
        \( d_2(\x^j,\ba^i) \)        & squared Euclidean distance between $\x^j \in \R^n$ and $\ba^i\in \R^n$ \\
        \( r(\ba^i) \)        & squared Euclidean distance between $\ba^i$ and its cluster center\\
        \bottomrule
    \end{tabular}
\end{table}

In cluster analysis, we consider a finite set $A$ of points in the $n$-dimensional space $\R^n$. In other words,
$$A = \{ \ba^1, \ldots, \ba^m \},\quad \text{where } \ba^i \in \R^n, \; i = 1, \ldots, m.$$
Each data point $\ba^i$ has $n$ \emph{features}.

The \emph{hard unconstrained clustering problem} entails partitioning the points in the set $A$ into $k$ disjoint subsets $A^j$, $j = 1, \ldots, k$, based on specific predefined criteria such that
\begin{enumerate}
    \item $A^j \neq \emptyset$ for all $j=1,\ldots,k$;
    \item $A^j \bigcap A^l = \emptyset$ for all $j,l = 1,\ldots,k, \text{ } j\neq l$; and
    \item $A = \bigcup_{j = 1}^k A^j$.
\end{enumerate}
The subsets $A^j$, $j = 1, \ldots, k$, are referred to as \emph{clusters}. Each cluster $A^j$ can be characterized by its \emph{center} $\x^j \in \R^n$, $j = 1, \ldots, k$. The task of determining these centers is known as the \emph{k-clustering problem} \cite{BagKarTah:2025}.

The concept of a similarity measure is essential for formulating the clustering problem. Typically, a similarity measure is defined utilizing a distance metric. In this case, we define it using the squared Euclidean norm (the $L_2$-norm) as the distance
$$
d_2(\x, \ba)=\|\x-\ba\|^2= \sum_{i=1}^{n} (x_i - a_i)^2,
$$
where $\x, \ba \in \R^n$.

When the cluster centers are known, each data point is assigned to the cluster with the nearest center. Then, the \emph{cluster function for the cluster} $A^j, \; j=1,\ldots,k$, is defined as
\begin{align}
\label{funk_klustereille}
    f_{A^j}(\x) = \sum_{\ba^i \in A^j} d_2(\x, \ba^i),
\end{align}
where $\x \in \R^n$ is treated as the center of the cluster $j$. The cluster function \eqref{funk_klustereille} is subsequently used to determine which cluster is selected for splitting.

\textsc{Clust-Splitter} is an incremental algorithm, meaning that it builds clusters incrementally. In this approach, the solution to the $(k-1)$-clustering problem is used to produce good starting points for the $k$-clustering problem. Suppose we are at the iteration $k$. First, we define the clusters $A^j$, $j=1,\ldots,k-1$, based on the solution to the previous $(k-1)$-clustering problem. Note that the case with $k=1$ is a convex optimization problem, which is straightforward to solve. We select the cluster with the highest cluster function value \eqref{funk_klustereille} and denote the index of this cluster by $c$. We split the cluster $A^c$, and the two newly formed clusters, together with the solution to the $(k-1)$-clustering problem, are used to generate a starting point for the $k$-clustering problem. Finally, we solve the $k$-clustering problem. The process is repeated until the desired number of clusters is reached. The three different clustering problems utilized by the new \textsc{Clust-Splitter} method are introduced in the following subsections. These problems are: the starting point auxiliary problem, the 2-clustering auxiliary problem and the $k$-clustering problem.

\subsection{Starting point auxiliary problem}

With the help of the \emph{starting point auxiliary} (SPA) problem, the aim is to identify two appropriate starting cluster centers before the selected cluster can be split into two. While one of the centers is the center of the splitted cluster, the other is determined through the solution of the SPA problem. This ensures that the splitting process of the selected cluster begins with a well-chosen starting point, which is crucial for the success of the clustering algorithm.

Assume that $c \in \{1,\ldots,k\}$ is the index of the cluster being split and the center of cluster $A^c$ is $\tilde{\x}^c \in \R^n$. First, we define
$$ r(\ba^i) = d_2(\tilde{\x}^c,\ba^i), \quad \text {where } \; \ba^i \in A^c.$$ Therefore, $r(\ba^i)$ represents the squared distance between the cluster center $\tilde{\x}^c$ and the data point $\ba^i$.
Second, the \emph{SPA function} is defined as
\begin{align}
\label{funk_apu1}
    \tilde{f}(\z)= \sum_{\ba^i \in A^c} \min \left\{ r(\ba^i), d_2(\z, \ba^i) \right\},
\end{align}
where $\z \in \R^n$. Note that $\tilde{f}(\z) \leq \sum_{\ba^i \in A^c} r(\ba^i) = f_{A^c}(\tilde{\x}^c)$ for all $\z \in \R^n$, and the inequality is strict if the distance from any data point to $\z$ is smaller than to $\tilde{\x}^c$. The function $\tilde{f}$ is nonsmooth, LLC and typically nonconvex as a sum of minima of convex functions. Finally, the \emph{SPA problem} is formulated as
\begin{align}
\label{teht_apu1}
\begin{cases}
\text{minimize} \quad& \tilde{f}(\z) \\
\text{subject to} \quad& \z \in \R^{n}.
\end{cases}
\end{align}

\subsection{2-clustering auxiliary problem}

The 2-clustering auxiliary problem focuses on splitting the selected cluster into two separate clusters. After solving the 2-clustering auxiliary problem, we have finished the splitting of the considered cluster and produced the centers of the two newly formed clusters. Assume that $c \in \{1,\ldots,k\}$ is the index of the cluster being split. The \emph{2-clustering auxiliary function} is defined as
$$ \hat{f}(\y)= \sum_{\ba^i \in A^c} \min\left\{d_2(\y^1, \ba^i), d_2(\y^2, \ba^i) \right\},$$
where $\y = (\y^1,\y^2) \in \R^{2n}$. The function $\hat{f}$ is also nonsmooth, LLC and typically nonconvex as a sum of minima of convex functions.
The \emph{2-clustering auxiliary problem} is formulated as
\begin{align}
\label{teht_apu2}
\begin{cases}
\text{minimize} \quad& \hat{f}(\y) \\
\text{subject to} \quad& \y = (\y^1,\y^2) \in \R^{2n}.
\end{cases}
\end{align}

\subsection{$k$-clustering problem}

The $k$-clustering problem is the main clustering task. As a result of solving the $k$-clustering problem, the process yields the centers of $k$ clusters. The \emph{$k$-clustering function}, or \emph{$k$th cluster function}, is defined as
\begin{align}
\label{funk_k_klusteria}
    f_k(\x)= \sum_{i=1}^{m} \min_{j=1,\ldots,k} d_2(\x^j, \ba^i),
\end{align}
where $\x = (\x^1, \ldots, \x^k) \in \R^{nk}$. The function $f_k$ is LLC for any $k$, convex for $k=1$, and nearly always nonconvex and nonsmooth for $k > 1$.
The \emph{$k$-clustering problem} is formulated as
\begin{align}
\label{teht}
\begin{cases}
\text{minimize} \quad& f_k(\x) \\
\text{subject to} \quad& \x = (\x^1, \ldots, \x^k) \in \R^{nk}.
\end{cases}
\end{align}

\section{\textsc{Clust-Splitter} method} \label{kappale4}

In this section, we introduce the new \textsc{Clust-Splitter} method, an incremental algorithm designed to solve MSSC problems. Suppose that we have already solved the $(k-1)$-clustering problem and are now moving on to solve the $k$-clustering problem. First, auxiliary problems are used to split a selected cluster from the previous solution into two smaller clusters. The resulting configuration, together with the previous solution of the $(k-1)$-clustering problem, provides a foundation (i.e., a starting point) for solving the $k$-clustering problem on the entire dataset. This process is repeated until the user-specified maximum number of clusters, $k_{max}$, is achieved. Therefore, \textsc{Clust-Splitter} solves not only the $k_{max}$-clustering problem but also all intermediate $k$-clustering problems for $k=1,\ldots, k_{max}-1$. \textsc{Clust-Splitter} uses the LMBM \cite{Haa:2004, HaaMieMak:2004, HaaMieMak:2007} to solve all optimization problems.

\begin{figure} [ht!]
    \centering
    \includegraphics[scale=0.46]{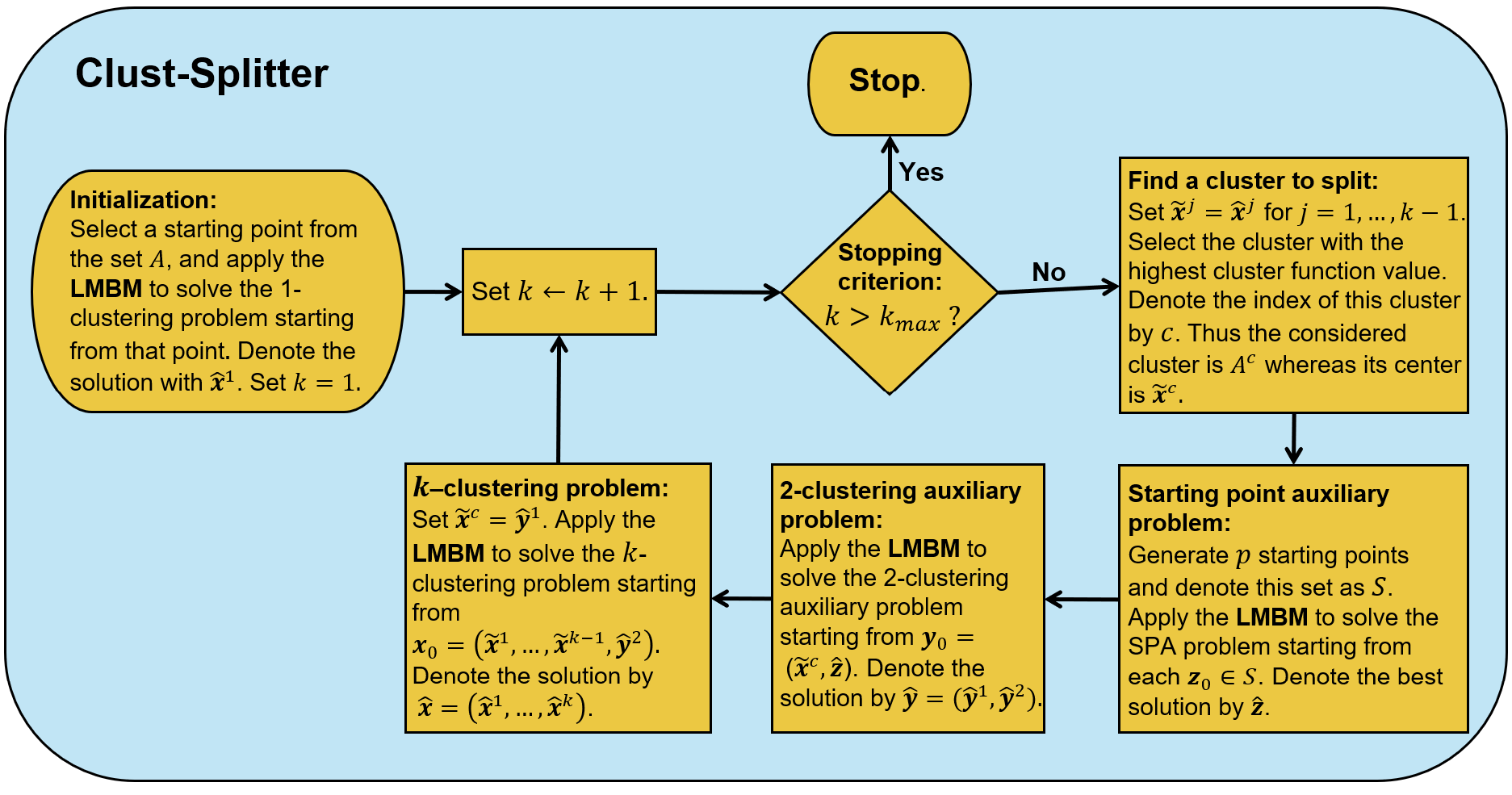}
    \caption{The \textsc{Clust-Splitter} method. Note that if $k=2$, after solving the 2-clustering auxiliary problem, we set $\hat{\x} = (\hat{\y}^1,\hat{\y}^2)$ and skip the step with the $k$-clustering problem.}
    \label{kuva_kaavio}
\end{figure}

A key aspect of \textsc{Clust-Splitter} is the strategic use of cluster splitting, which facilitates the identification of high-quality starting points, an essential factor in achieving global or near-global solutions in nonconvex clustering problems. This is also a notable advantage of the new approach compared to the traditional selection of starting points and their corresponding auxiliary problems in incremental clustering algorithms (see, e.g., \cite{BagTahUgo:2016, KarBagTah:2018, BigClust:2025}). Namely, the auxiliary problems in \textsc{Clust-Splitter} are smaller and therefore significantly easier to solve. Furthermore, in the method, additional criteria for selecting the cluster to be split can be adjusted using parameters. For instance, splitting can be prohibited for clusters containing fewer than five data points (i.e., outlier clusters). Figure \ref{kuva_kaavio} illustrates the structure of the \textsc{Clust-Splitter} method, and the exact algorithm is presented in Algorithm \ref{Clust-Splitter_pseudo}. Next, we take a closer look at the three main parts of the \textsc{Clust-Splitter} method. In each part, one of the optimization problems \eqref{teht_apu1}, \eqref{teht_apu2} or \eqref{teht} is solved.

Suppose that we have just started the iteration $k$. First, we define the clusters $A^j$, $j=1,\ldots,k-1$, based on the solution $\hat{\x}=(\hat{\x}^1,\ldots,\hat{\x}^{k-1})$ to the previous $(k-1)$-clustering problem. Second, we store this previous solution to $\tilde{\x}=(\tilde{\x}^1,\ldots,\tilde{\x}^{k-1})$. Third, we calculate the cluster function value $f_{A^j}(\tilde{\x}^j)$ using \eqref{funk_klustereille} for each cluster $A^j, \; j=1,\ldots,k-1$, and select the one with the highest value for splitting. We denote the index of this cluster by $c$. Thus, in the splitting part of the method, the considered cluster is $A^c$ whereas its center is $\tilde{\x}^c$. At this point, we are ready to proceed with the auxiliary problems.

The SPA problem \eqref{teht_apu1} in Step 3 of Algorithm \ref{Clust-Splitter_pseudo} aims to identify the best possible starting point for the 2-clustering auxiliary problem \eqref{teht_apu2} in order to split the cluster $A^c$. The SPA problem is computed $p$ times, where the number $p$ is specified by the user. Each of these runs use a different starting point, which can be chosen, for example, as a center of randomly selected data points from the splitted cluster $A^c$. For a more detailed description, see Remark \ref{remark_startingpoints}. The solution that yields the smallest starting point auxiliary function \eqref{funk_apu1} value is denoted by $\hat{\z}$.

The goal of the 2-clustering auxiliary problem \eqref{teht_apu2} in Step 4 of Algorithm \ref{Clust-Splitter_pseudo} is to split the cluster $A^c$ into two parts, resulting in the centers $\hat{\y}^1$ and $\hat{\y}^2$ of the newly formed clusters. The starting point for the problem \eqref{teht_apu2} consists of the center $\x^c$ of the considered cluster $A^c$ and the additional center $\hat{\z}$ determined by the SPA problem \eqref{teht_apu1}. In other words, the starting point is $\y_0=(\tilde{\x}^c, \hat{\z})$. The solution of the 2-clustering auxiliary problem consists of the centers of the two new clusters, stored in the vector $\hat{\y} = (\hat{\y}^1, \hat{\y}^2)$.

After solving the two auxiliary problems, we are finally ready to address the main problem: the $k$-clustering problem \eqref{teht} in Step 5 of Algorithm \ref{Clust-Splitter_pseudo}. The input to the $k$-clustering problem is the solution $\hat{\y}=(\hat{\y}^1,\hat{\y}^2)$ to the 2-clustering auxiliary problem splitting the cluster $A^c$. Using this input we construct the starting point for the $k$-clustering problem as 
$$\x_0 = (\x_0^1, \ldots, \x_0^k) = (\tilde{\x}^1, \ldots, \tilde{\x}^{k-1}, \hat{\y}^2),$$
where $\tilde{\x}^1, \ldots, \tilde{\x}^{k-1}$ are otherwise the cluster centers obtained from the $(k-1)$-clustering problem \eqref{teht} but $\tilde{\x}^c=\hat{\y}^1$. The solution to the $k$-clustering problem consists of the updated cluster centers and is denoted by $\hat{\x} = (\hat{\x}^1, \ldots, \hat{\x}^k)$. For example, if $k=4$ then we know three cluster centers $\tilde{\x}^1$, $\tilde{\x}^2$ and $\tilde{\x}^3$ from the previous iteration of the method. Then if the second cluster is being split, after obtaining $\hat{\y}^1$ and $\hat{\y}^2$, the starting point of the 4-clustering problem will be $\x_0 = (\tilde{\x}^1,\hat{\y}^1,\tilde{\x}^3,\hat{\y}^2)$.

\begin{algoritmi} {\textsc{\textbf{Clust-Splitter}}}
\noindent
\begin{itemize}[leftmargin=1.85cm]
\label{Clust-Splitter_pseudo}

\item[\textbf{Input:}]
    The dataset $A$, the maximum number of clusters $k_{max} \geq 1$, and the number of starting points $p \geq 1$ for the auxiliary problem \eqref{teht_apu1}.
    
\item[\textbf{Output:}]
    The solution $\hat{\x}$ to the $k_{max}$-clustering problem and all the intermediate solutions to the $k$-clustering problems with $k=1,\ldots,k_{max}-1$.

\item[\textbf{Step 0.}]
    \emph{Initialization:}
    Select a starting point from the set $A$, and apply the LMBM to solve the 1-clustering problem \eqref{teht} starting from that point. Denote the solution with $\hat{\x}^1$. Set $k=1$.

\item[\textbf{Step 1.}]
    \emph{Stopping criterion:}
    Set $k \leftarrow k+1$. If $k>k_{max}$, STOP: the $k_{max}$-clustering problem has been solved.

\item[\textbf{Step 2.}]
    \emph{Find a cluster to split:}
    Define the clusters $A^j$, $j=1,\ldots,k-1$, based on the solution to the previous $(k-1)$-clustering problem. Set $\tilde{\x}^j = \hat{\x}^j$ for $j=1,\ldots,k-1$. Calculate the cluster function value $f_{A^j}(\tilde{\x}^j)$ using \eqref{funk_klustereille} for each cluster $A^j, \; j=1,\ldots,k-1$, and select the one with the highest value. Denote the index of this cluster by $c$. Thus, the considered cluster is $A^c$ whereas its center is $\tilde{\x}^c$.

\item[\textbf{Step 3.}]
    \emph{Starting point auxiliary problem:}
    Generate $p$ starting points from $\R^n$ and denote this set as $S$. Apply the LMBM to solve the SPA problem \eqref{teht_apu1} starting from each $\z_0 \in S$. Choose the solution that gives the smallest value of the auxiliary function \eqref{funk_apu1} and denote it by $\hat{\z}$. 

\item[\textbf{Step 4.}]
    \emph{2-clustering auxiliary problem:}
    Apply the LMBM to solve the 2-clustering auxiliary problem \eqref{teht_apu2} starting from $\y_0 = (\tilde{\x}^c, \hat{\z})$. Denote the solution by $\hat{\y} = (\hat{\y}^1,\hat{\y}^2)$. If $k=2$, set $\hat{\x} = (\hat{\y}^1,\hat{\y}^2)$ and go to Step 1.

\item[\textbf{Step 5.}]
    \emph{$k$-clustering problem:}
    Set $\tilde{\x}^c = \hat{\y}^1$. Apply the LMBM to solve the $k$-clustering problem \eqref{teht} starting from $\x_0 = (\tilde{\x}^1,\ldots,\tilde{\x}^{k-1},\hat{\y}^2)$. Denote the solution by $\hat{\x} = (\hat{\x}^1,\ldots,\hat{\x}^k)$. Go to Step 1.

\end{itemize}
\end{algoritmi}

\begin{remark}
    At Step 0 of Algorithm \ref{Clust-Splitter_pseudo}, the starting point is defined as the center of $M_1 > 1$ randomly selected data points from $A$. Note that here only a single starting point is sufficient, as the 1-clustering problem is convex.
\end{remark}

\begin{remark}
\label{remark_startingpoints}
In Step 3 of Algorithm \ref{Clust-Splitter_pseudo}, the starting points for the auxiliary problem \eqref{teht_apu1} can be defined, for example, as follows:
\begin{itemize}
    \item One or more starting points are generated by computing the center of $M_1 > 1$ randomly selected data points from the cluster $A^c$ being split.
    \item One or more starting points are computed as the center of $M_2 > 1$ randomly selected data points from the cluster $A^c$ being split, with the additional requirement that each candidate center must be sufficiently distant from the current center of the cluster $A^c$. If this condition is not met, a new center is generated using another set of $M_2$ random data points.
    \item One starting point is selected as the current center of the cluster $A^c$ being split.
\end{itemize}
\end{remark}

As we have seen, the LMBM is used to solve all the optimization problems encountered in \textsc{Clust-Splitter}. This method is developed to handle general large-scale nonconvex and nonsmooth optimization problems. For simplicity, we assume in the following that the dimension of the considered problem is $N$. The core idea behind the LMBM is to generate search directions using a variable metric matrix updated with classical BFGS and SR1 formulas, which preserve matrix sparsity. However, instead of storing the full matrix, the LMBM maintains only a limited number of so-called correction vectors. These vectors are used to compute the search direction as a product of the matrix and a subgradient, which significantly improves computational efficiency. In addition, the LMBM requires that the function value $f(\x)$ and an arbitrary subgradient $\boldsymbol{\xi}$ from the subdifferential $\partial f(\x)$ can be computed at any point $\x \in \R^N$. This requirement is easily met for all of our three problems, namely \eqref{teht_apu1}, \eqref{teht_apu2}, and \eqref{teht}. For further details on the LMBM, see \cite{HaaMieMak:2004}.

The convergence of the LMBM has been proven in \cite{HaaMieMak:2007}. Next, we present the three assumptions required for the convergence.

\begin{assumption}
The objective function $f: \R^N \to \R$ is LLC.
\end{assumption}

\begin{assumption}
The objective function $f: \R^N \to \R$ is upper semi-smooth \cite{bih:1984}.
\end{assumption}

\begin{assumption}
The level set $\{\x \in \R^N \; | \; f(\x) \leq f(\x_0)\}$ is bounded for every starting point $\x_0 \in \R^N$.
\end{assumption}

\noindent These three assumptions are trivially satisfied for the SPA problem, the 2-clustering auxiliary problem, and the $k$-clustering problem. Since under the above assumptions, the following convergence theorems hold, these results apply to each problem solved during \textsc{Clust-Splitter}.

\begin{theorem}
\cite{HaaMieMak:2007} If the LMBM terminates after a finite number of iterations, at iteration $h$, then the point $\x_h$ is a stationary point of the problem.
\end{theorem}

\begin{theorem}
\cite{HaaMieMak:2007} Every accumulation point produced by the LMBM is a stationary point of the problem.
\end{theorem}

\section{Numerical experiments} \label{kappale5}

To demonstrate the performance of the proposed \textsc{Clust-Splitter} method, we compare it with five other clustering algorithms: \textsc{LMBM-Clust} \cite{KarBagTah:2018}, \textsc{DC-Clust} \cite{BagTahUgo:2016}, \textsc{Big-Clust} \cite{BigClust:2025}, \textsc{Big-Means} \cite{MusMlaJarMus2023}, and \textsc{MiniBatchKMeans} \cite{scikit}. \textsc{LMBM-Clust}, \textsc{DC-Clust}, and \textsc{Big-Clust} are incremental clustering methods, with \textsc{LMBM-Clust} utilizing the LMBM, \textsc{DC-Clust} leveraging the nonsmooth DC-representation of the MSSC problem, and \textsc{Big-Clust} employing stochastic optimization. \textsc{Big-Means} and \textsc{MiniBatchKMeans} are variants of the traditional $k$-means algorithm, with \textsc{Big-Means} tailored for large-scale data and \textsc{MiniBatchKMeans} using a mini-batch optimization approach. The tests are conducted on large-scale real-world data, as well as through external validation using small real-world and simulated datasets.

The solvers \textsc{Clust-Splitter}, \textsc{LMBM-Clust}, \textsc{DC-Clust}, and \textsc{Big-Clust} are implemented in Fortran 95, while the solvers \textsc{Big-Means} and \textsc{MiniBatchKMeans} are implemented in Python. All computational experiments are carried out on an Intel(R) Core(TM) i3-1215U CPU (1.20GHz, 4.40GHz) running under Windows 10. We use gfortran to compile the Fortran codes and Python 3.9.7 for Python codes with NumPy 1.26.4 and Numba 0.60.0 for \textsc{Big-Means}. Additionally, we follow the provided implementations and adhere to the recommended default parameter values for all methods, as specified in their respective references. An open-source implementation of the proposed \textsc{Clust-Splitter} algorithm is given at \url{https://github.com/jmlamp/Clust-Splitter}. Additionally, in \textsc{Clust-Splitter}, the number of starting points is set to $p=3$ and we generate one starting point using each of the ways described in Remark \ref{remark_startingpoints}, where the numbers of random data points are set to $M_1 = 10$ and $M_2 = 7$.

\subsection{Experiments in large-scale real-world data}

We conducted our experiments on a collection of 18 publicly available large-scale real-world datasets. They are the same as in \cite{BigClust:2025}, but four datasets were excluded from the analysis due to computer memory limitations. Summary information for the datasets used is presented in Table \ref{taulukko_datat}, while more detailed descriptions are accessible via the references provided. All datasets consist exclusively of numeric features, with no missing values. They also vary in size, with the number of features ($n$) ranging from 2 to 5000 and the number of records ($m$) from 7797 to 581,012.

\begin{table}[ht!]
\centering
\caption{Brief description of the datasets used.} \vspace{0.25cm}
\label{taulukko_datat}
\begin{tabular}{lllll}
\toprule
dataset & $m$ & $n$ & $m \times n$ & References \\
\midrule
ISOLET & 7797 & 617 & 4,810,749 & \cite{isolet} \\
Gisette & 13,500 & 5000 & 67,500,000 & \cite{Gisette} \\
Gas Sensor Array Drift & 13,910 & 128 & 1,780,480 & \cite{GasSensor} \\
EEG Eye State & 14,980 & 14 & 209,720 & \cite{eeg_eye_state} \\
D15112 & 15,112 & 2 & 30,224 & \cite{tsplib} \\
Online News Popularity & 39,644 & 58 & 2,299,352 & \cite{OnlineNewsPopularity} \\
KEGG Metabolic & 53,413 & 20 & 1,068,260 & \cite{KEGGMetabolic} \\
Shuttle Control & 58,000 & 9 & 522,000 & \cite{uci} \\
Sensorless Drive Diagnosis & 58,509 & 48 & 2,808,432 & \cite{sensorless} \\
MFCCs for Speech Emotion Recognition & 85,134 & 58 & 4,937,772 & \cite{data:mfcc} \\
Pla85900 & 85,900 & 2 & 171,800 & \cite{tsplib} \\
Music Analysis & 106,574 & 518 & 55,205,332 & \cite{music} \\
MiniBooNE Particle Identification & 130,064 & 50 & 6,503,200 & \cite{miniboone} \\
Protein Homology & 145,751 & 74 & 10,785,574 & \cite{data:protein} \\
Range Queries Aggregates & 200,000 & 7 & 1,400,000 & \cite{RangeQueries} \\
Skin Segmentation & 245,057 & 3 & 735,171 & \cite{SkinSegmentation} \\
3D Road Network & 434,874 & 3 & 1,304,622 & \cite{3DRoadNetwork} \\
Covertype & 581,012 & 10 & 5,810,120 & \cite{covertype} \\
\bottomrule
\end{tabular}
\end{table}

\subsubsection{Setup overview}

\textsc{Clust-Splitter}, \textsc{LMBM-Clust}, \textsc{DC-Clust}, and \textsc{Big-Clust} utilize an incremental approach to solve clustering problems. Using these methods, we incrementally compute up to 25 clusters in each dataset. In contrast, since \textsc{Big-Means} and \textsc{MiniBatchKMeans} do not produce intermediate results, we conduct multiple runs in each dataset with different numbers of clusters (denoted as $k$) for comparison purposes. Due to the stochastic characteristics, \textsc{Big-Clust}, \textsc{Big-Means}, and \textsc{MiniBatchKMeans} are executed ten times, with the results averaged over these runs. In other words, each dataset with $k$ clusters is computed ten times, and the results are averaged. The incremental algorithm \textsc{Big-Clust} computes up to 25 clusters in a single run, and the algorithm is executed 10 times for each dataset. For \textsc{Big-Means}, each dataset with $k$ clusters needs ten separate runs, whereas \textsc{MiniBatchKMeans} is able to produce these ten runs in a single loop. The CPU time for all algorithms is limited to 20 hours per dataset, including all calculations up to 25 clusters.

We have employed the following metrics and parameters to compare the performance of the algorithms:
\begin{enumerate}
    \item \textbf{Cluster function values:} These are also referred to as the sum of squared errors (SSE), a prototype-based cohesion measure evaluating the average variation within clusters. Tables \ref{taulukko_isolet}\,--\,\ref{taulukko_covertype} in the Appendix include the best-known cluster function \eqref{funk_k_klusteria} value, $f_{best}$ (scaled by the dataset-specific multiplier shown after the name of the dataset), and the relative errors for each algorithm. The relative error $E_k$ for a dataset with $k$ clusters is calculated as
    $$ E_k = \frac{f_k - f_{best}}{f_{best}} \times 100 \%, $$
    where $f_k$ is the $k$-clustering function \eqref{funk_k_klusteria} value obtained by the algorithm. We adopt the $f_{best}$ values reported in \cite{BigClust:2025} unless better results are achieved in our experiments. In this study, these better values are marked with an asterisk. In addition, we give the average of relative errors $E_\text{aver}$ for each method, and it is calculated as $E_\text{aver} = \frac{1}{8} \sum_{k \in K} E_k$, where $K = \{ 2, 3, 4, 5, 10, 15, 20, 25 \}$. 
    
    \item \textbf{Computational time (in seconds):} Tables \ref{taulukko_isolet}\,--\,\ref{taulukko_covertype} present the time measurements separately for different phases: the time required to read the data ($t_\text{init}$), the time taken to compute $k$ clusters ($t_k$), and the total computation time ($t_\text{total}$) for all clusters. For the \textsc{Clust-Splitter}, \textsc{LMBM-Clust}, \textsc{DC-Clust}, and \textsc{Big-Clust}, the total time is calculated as: $t_\text{total} = t_\text{init} + t_{25}$. In contrast, for \textsc{Big-Means} and \textsc{MiniBatchKMeans}, the total time follows the formula $t_\text{total} = t_\text{init} + \sum_{k \in K} t_k$, where $K = \{ 2, 3, 4, 5, 10, 15, 20, 25\}$.
    
    \item \textbf{Davies-Bouldin and Dunn cluster validity indices:} Definitions of these indices are given below.
\end{enumerate}

The validity indices are calculated as follows: Let $A^1, \ldots, A^k$ represent a cluster distribution of the set $A$, where the number of clusters $k > 1$ and $\x^1,\ldots,\x^k$ are the cluster centers of these clusters. Let $d(\x^i, \x^j) = \sqrt{d_2(\x^i, \x^j)}$ denote the Euclidean distance between cluster centers $\x^i$ and $\x^j$. The Davies-Bouldin index (DBI) is defined as \cite{DavBou:1979}
$$ \text{DBI} = \frac{1}{k} \sum_{i=1}^{k} \max_{j=1, \ldots, k, \; j \neq i} \frac{S_k(A^i) + S_k(A^j)}{d(\x^i, \x^j)}.$$
Here, $S_k(A^l)$ represents the average distance of all data points in the cluster $A^l$ to its corresponding cluster center $\x^l$. The DBI ratio becomes smaller when the clusters are well-separated and compact. Therefore, the optimal number of clusters minimizes the DBI value. The Dunn index (DI) is defined as \cite{Dun:1974}
$$
\text{DI} = 
\frac{
    \displaystyle \min_{\substack{i,j = 1, \ldots, k, \; i \neq j}} d(\x^i, \x^j)
}{
    \displaystyle \max_{l=1,\ldots,k} \left\{ \max_{\ba \in A^l} \|\x^l - \ba\| \right\}
},
$$
where the goal is to maximize inter cluster distances while minimizing intra cluster distances. Thus, the number of clusters that maximizes the DI can be regarded as the optimal number of clusters. In our study, these two indices are calculated only for the incremental algorithms \textsc{Clust-Splitter}, \textsc{LMBM-Clust}, \textsc{DC-Clust}, and \textsc{Big-Clust}.

\subsubsection{Results}

\begin{figure}[ht!]
    \centering
    \includegraphics[scale=1]{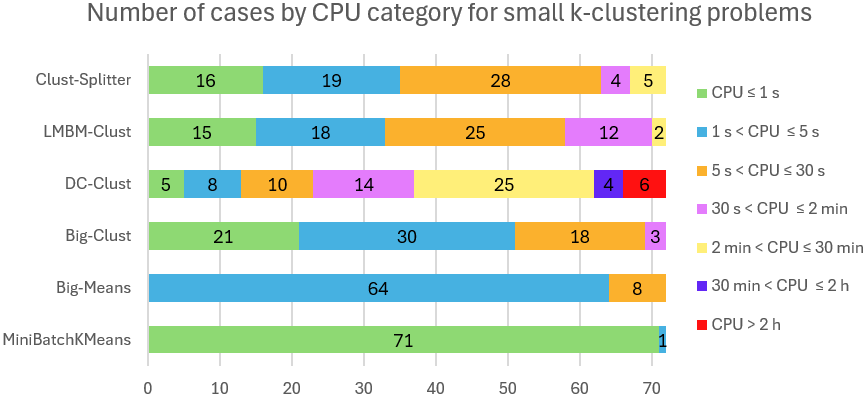}
    \caption{Cases categorized by CPU times for small $k$-clustering problems, where $k \leq 5$. The total number of cases per algorithm is 72, comprising 18 datasets and four different numbers of clusters $(k=2, 3, 4, 5)$.}
    \label{pienet_cpu}
\end{figure}

\begin{figure}[ht!]
    \centering
    \includegraphics[scale=1]{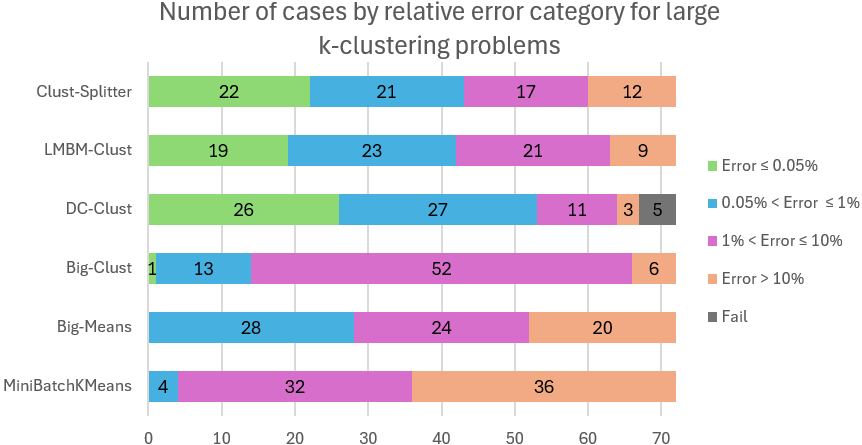}
    \caption{Cases categorized by relative errors for large $k$-clustering problems, where $k \geq 10$. The total number of cases per algorithm is 72, comprising 18 datasets and four different numbers of clusters $(k=10, 15, 20, 25)$. Note that if an algorithm fails to find a solution within the 20-hour time limit, the case is assigned to the category 'Fail'.}
    \label{isot_error}
\end{figure}

\begin{figure}[ht!]
    \centering
    \includegraphics[width=\textwidth]{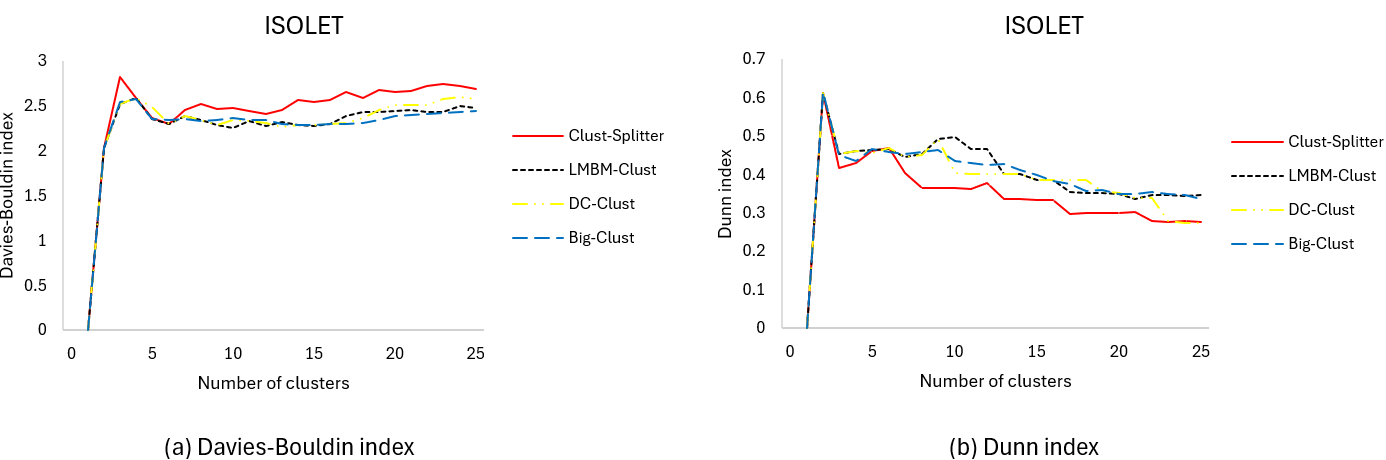}
    \caption{ISOLET: Davies-Bouldin and Dunn validity indices vs. number of clusters.}
    \label{index_isolet}
\end{figure}

\begin{figure}[ht!]
    \centering
    \includegraphics[width=\textwidth]{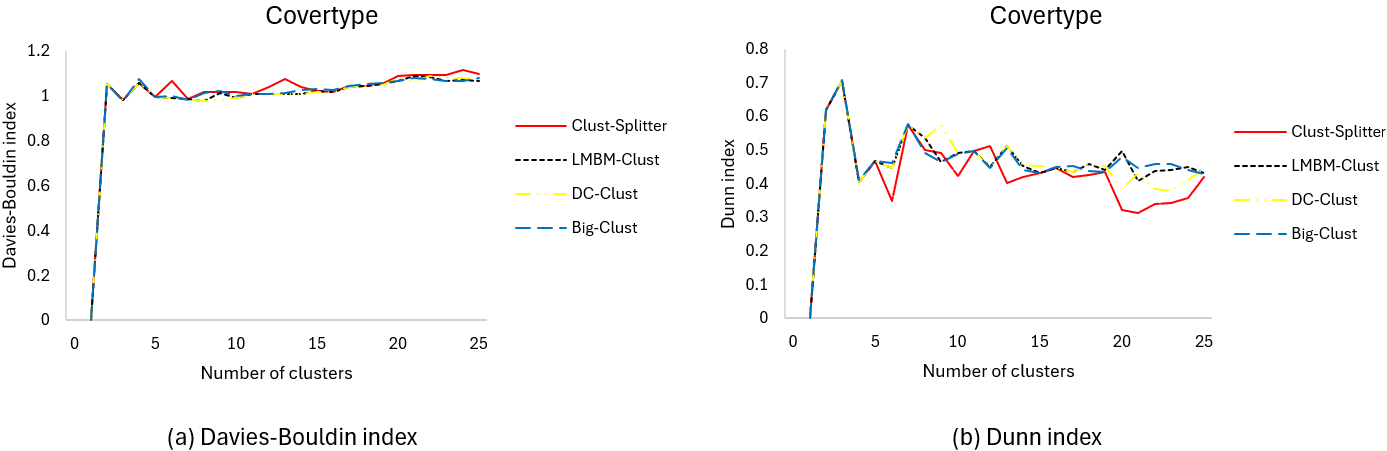}
    \caption{Covertype: Davies-Bouldin and Dunn validity indices vs. number of clusters.}
    \label{index_covertype}
\end{figure}

The results of the numerical experiments are summarized in Figures \ref{pienet_error}\,--\,\ref{isot_cpu}, with more detailed results provided in the Appendix: Tables \ref{taulukko_isolet}\,--\,\ref{taulukko_covertype} and Figures \ref{index_isolet}\,--\,\ref{index_covertype}. In the following, a \emph{case} refers to the problem of solving a dataset with a fixed number of clusters. Since we use different numbers of clusters, each dataset is thus divided into several cases. Figures \ref{pienet_error} and \ref{pienet_cpu} illustrate the number of cases classified by relative errors and CPU times for small $k$-clustering problems ($k=2,3,4,5$). Similarly, Figures \ref{isot_error} and \ref{isot_cpu} present the corresponding results for large $k$-clustering problems ($k=10,15,20,25$). As already mentioned, in the Appendix, Tables \ref{taulukko_isolet}\,--\,\ref{taulukko_covertype} present the detailed numerical results, including the best-known values of the $k$-clustering function, relative errors, and computational times for different algorithms. The validity indices for each dataset are summarized in Figures \ref{index_isolet}\,--\,\ref{index_covertype}, where part (a) represents the DBI values and part (b) corresponds to the DI values. Before proceeding with a detailed examination of the results, we first provide a general overview of the performance of each of the six methods.

\bigskip
\noindent \textbf{General overview of results}
\medskip

\begin{figure}[ht!]
    \centering
    \includegraphics[width=\textwidth]{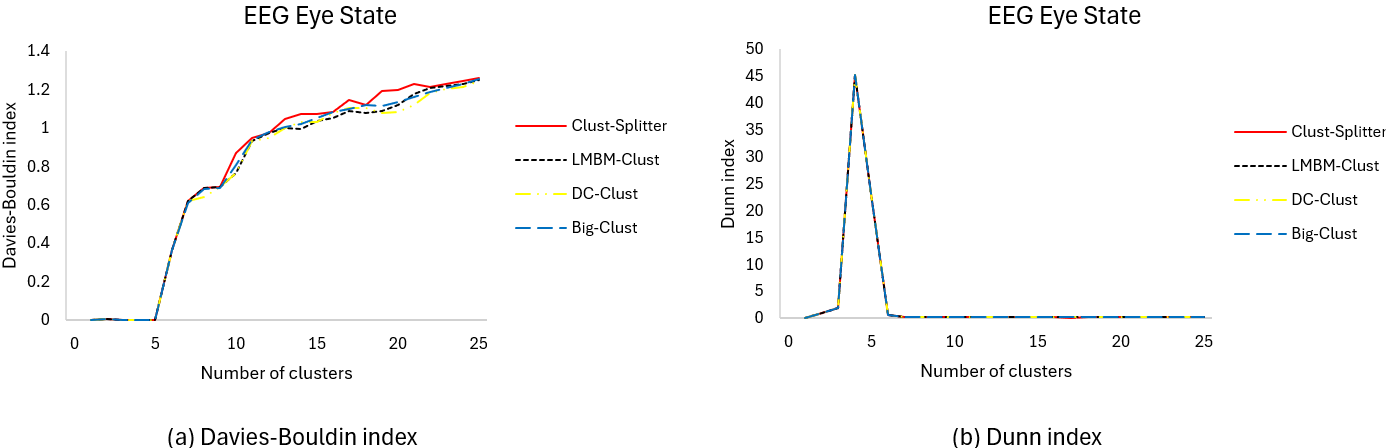}
    \caption{EEG Eye State: Davies-Bouldin and Dunn validity indices vs. number of clusters.}
    \label{index_eegeye}
\end{figure}

\begin{figure}[ht!]
    \centering
    \includegraphics[width=\textwidth]{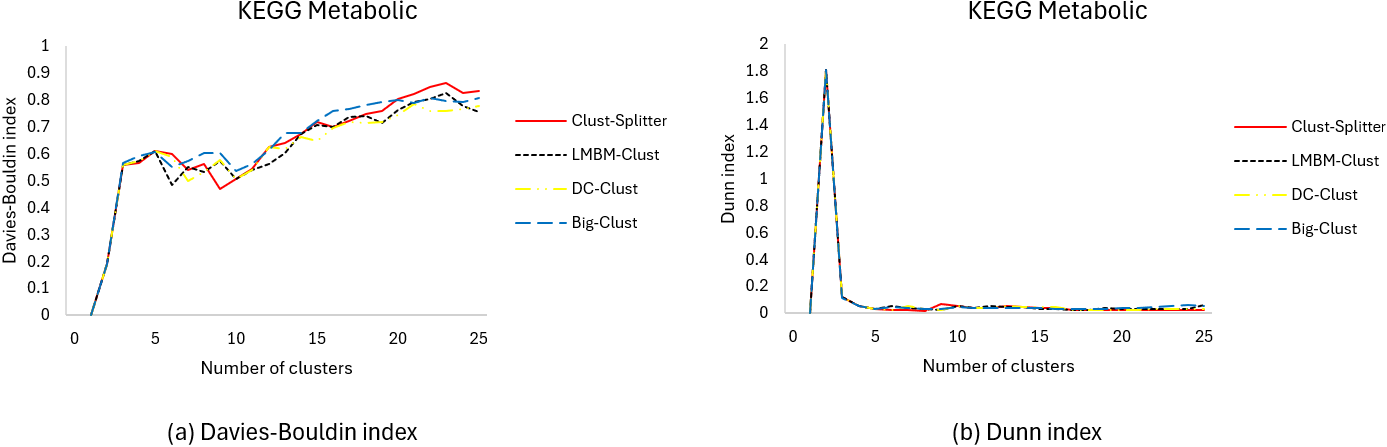}
    \caption{KEGG Metabolic: Davies-Bouldin and Dunn validity indices vs. number of clusters.}
    \label{index_kegg}
\end{figure}

\begin{figure}[ht!]
    \centering
    \includegraphics[width=\textwidth]{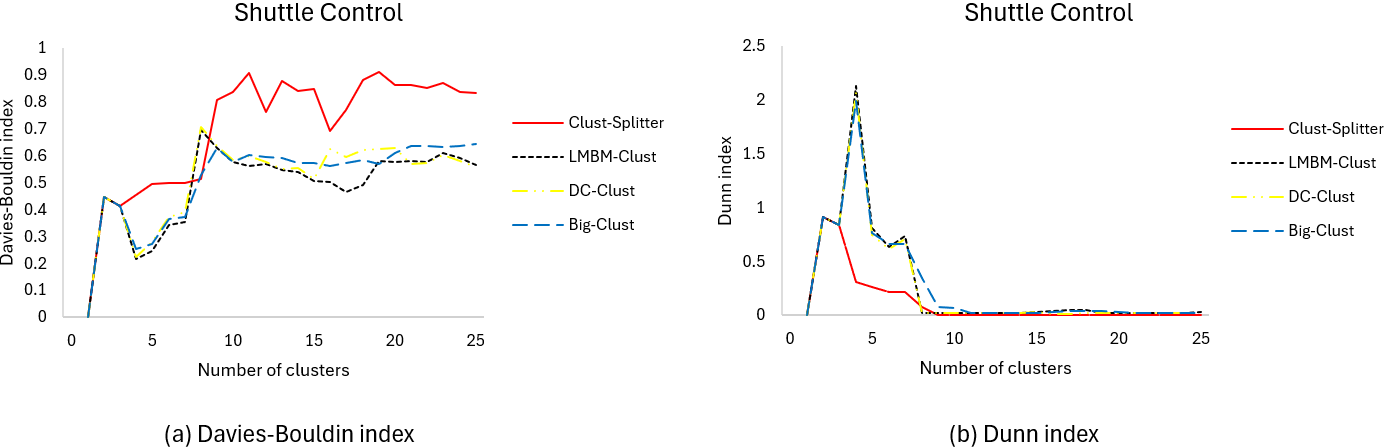}
    \caption{Shuttle Control: Davies-Bouldin and Dunn validity indices vs. number of clusters.}
    \label{index_shuttle}
\end{figure}

\noindent \textsc{MiniBatchKMeans} is clearly the fastest method, as shown in Figures \ref{pienet_cpu} and \ref{isot_cpu}. In fact, it is the fastest across all the datasets and the numbers of clusters (when comparing $t_k$ times), except for the dataset D15112 with $k=2$, where \textsc{Clust-Splitter} achieves the shortest run time (0.02 seconds for \textsc{Clust-Splitter} vs. 0.06 seconds for \textsc{MiniBatchKMeans}). Despite being the fastest among the analyzed methods, \textsc{MiniBatchKMeans} has relatively poor accuracy. It never achieves a relative error of 0.05\% or lower and exceeds a relative error of 10\% clearly more frequently than the other methods, for both small and large numbers of clusters (see Figures \ref{pienet_error} and \ref{isot_error}). Additionally, the performance of the method often declines as the number of clusters increases (see, e.g., Tables \ref{index_eegeye}, \ref{index_kegg}, and \ref{index_shuttle}).

\begin{figure}[ht!]
    \centering
    \includegraphics[scale=1]{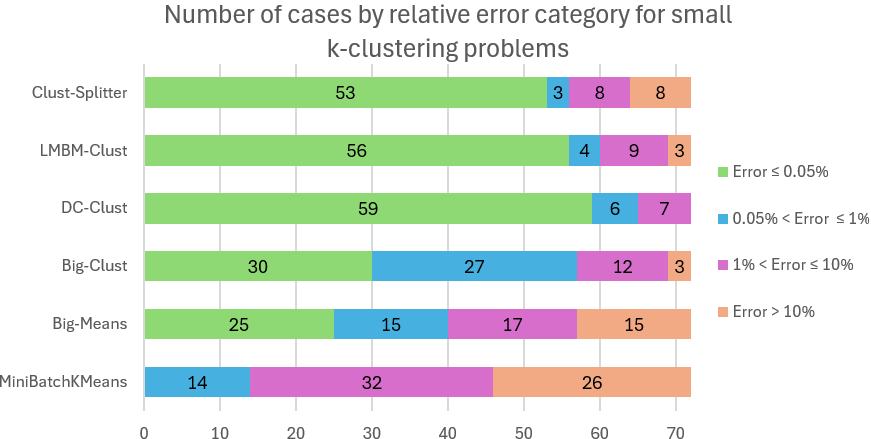}
    \caption{Cases categorized by relative errors for small $k$-clustering problems, where $k \leq 5$. The total number of cases per algorithm is 72, comprising 18 datasets and four different numbers of clusters $(k=2, 3, 4, 5)$.}
    \label{pienet_error}
\end{figure}

\begin{figure}[ht!]
    \centering
    \includegraphics[scale=1]{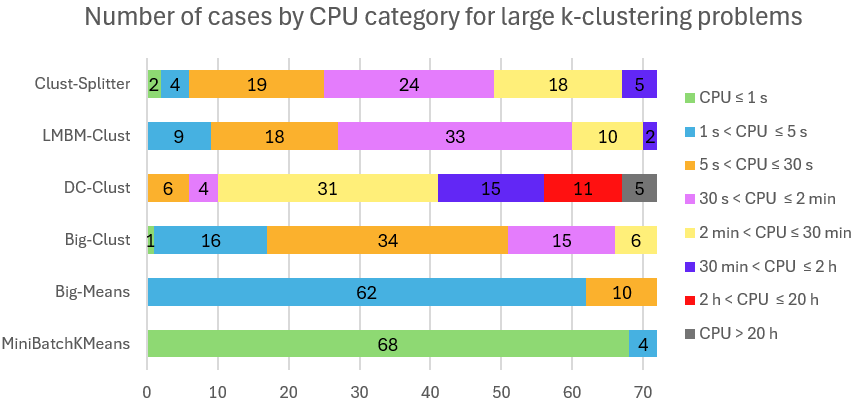}
    \caption{Cases categorized by CPU times for large $k$-clustering problems, where $k \geq 10$. The total number of cases per algorithm is 72, comprising 18 datasets and four different numbers of clusters $(k=10, 15, 20, 25)$. Note that the category 'CPU $>$ 20 h' denotes a failure to find a solution within the 20-hour time limit.}
    \label{isot_cpu}
\end{figure}

\begin{figure}[ht!]
    \centering
    \includegraphics[width=\textwidth]{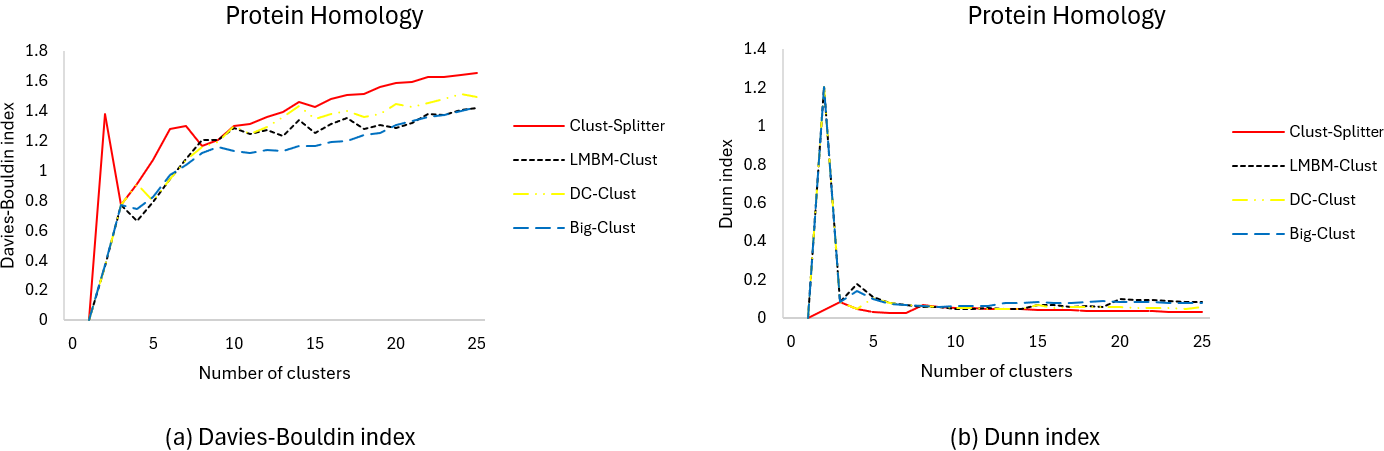}
    \caption{Protein Homology: Davies-Bouldin and Dunn validity indices vs. number of clusters.}
    \label{index_protein}
\end{figure}

\begin{figure}[ht!]
    \centering
    \includegraphics[width=\textwidth]{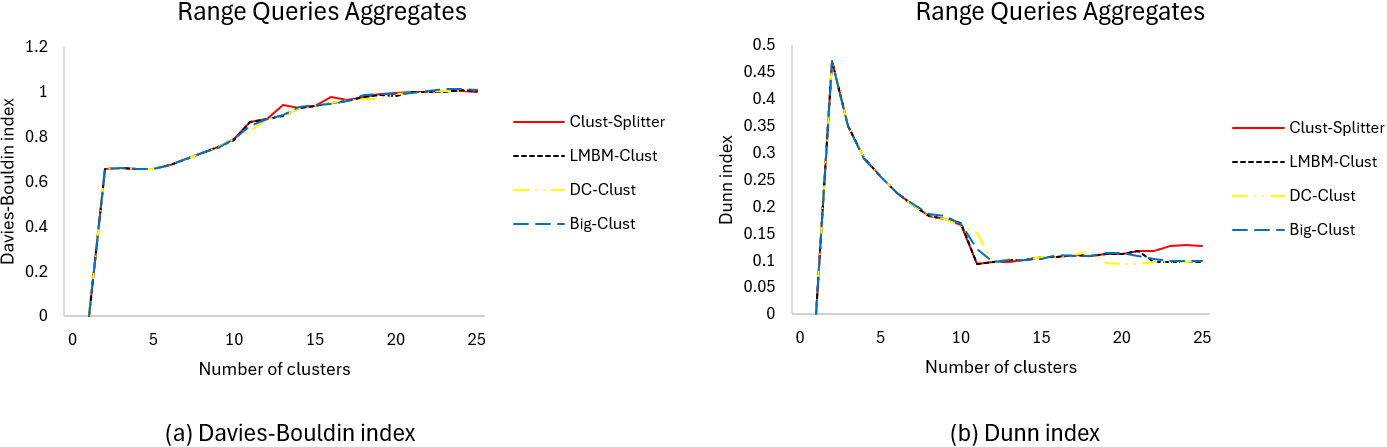}
    \caption{Range Queries Aggregates: Davies-Bouldin and Dunn validity indices vs. number of clusters.}
    \label{index_rangeque}
\end{figure}

Based on Figures \ref{pienet_cpu} and \ref{isot_cpu}, \textsc{Big-Means} is the second-fastest method. However, its accuracy is lower than that of the four incremental algorithms (see Figures \ref{pienet_error} and \ref{isot_error}, and, e.g., Tables \ref{index_eegeye} and \ref{index_kegg}). Although its relative errors $E_k$ are not as large as those of \textsc{MiniBatchKMeans}, \textsc{Big-Means} consistently produces errors across all numbers of clusters (see, e.g., Tables \ref{index_protein} and \ref{index_rangeque}). Notably, it finds the best known solution only in three cases, when $k=2$ (for Music Analysis, Skin Segmentation, and Covertype datasets).

\begin{figure}[ht!]
    \centering
    \includegraphics[width=\textwidth]{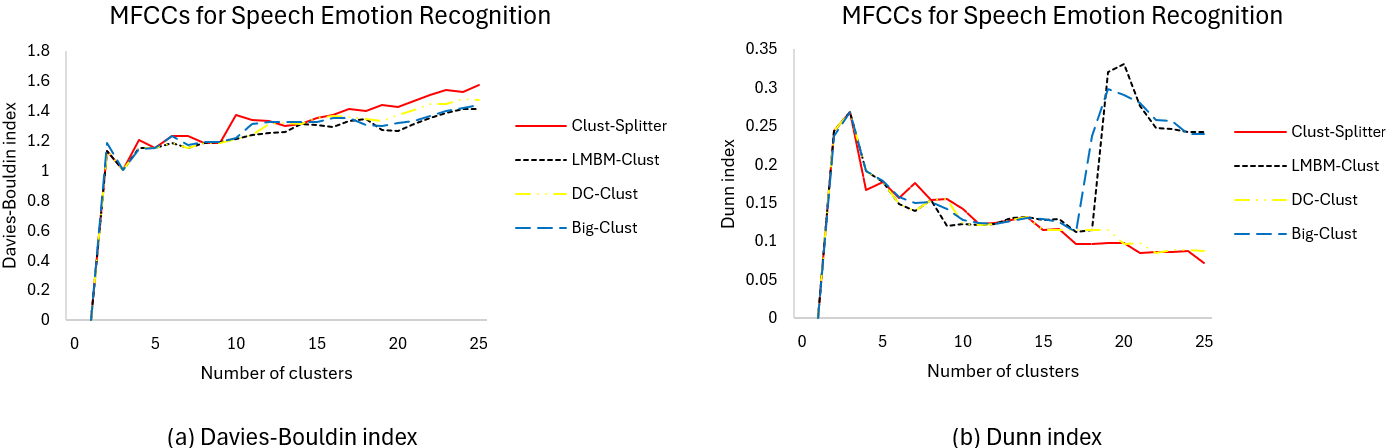}
    \caption{MFCCs for Speech Emotion Recognition: Davies-Bouldin and Dunn validity indices vs. number of clusters.}
    \label{index_mfcc}
\end{figure}

\begin{figure}[ht!]
    \centering
    \includegraphics[width=\textwidth]{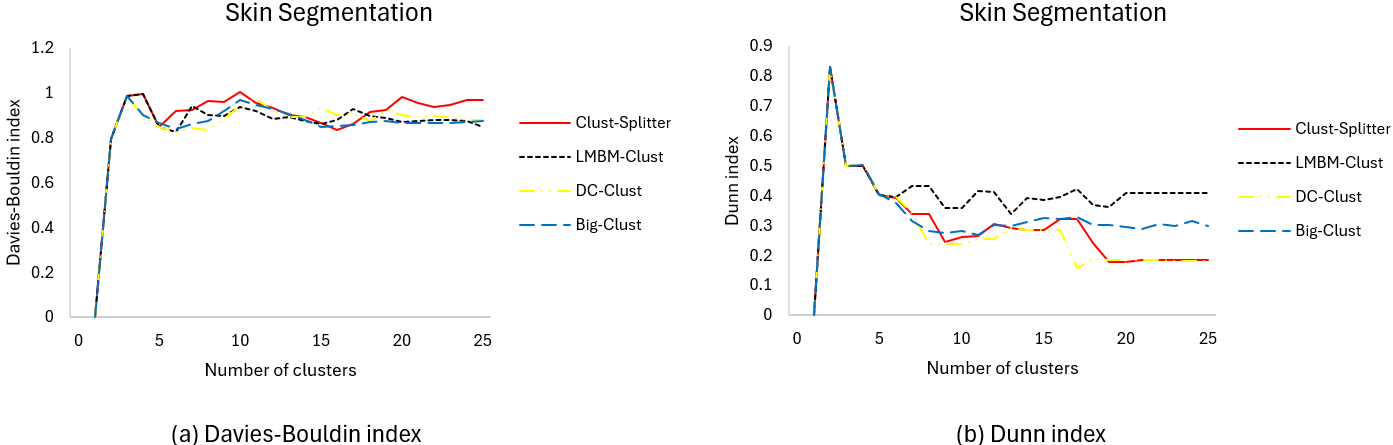}
    \caption{Skin Segmentation: Davies-Bouldin and Dunn validity indices vs. number of clusters.}
    \label{index_skin}
\end{figure}

Among the four incremental algorithms, \textsc{Big-Clust} has the highest relative errors (see Figures \ref{pienet_error} and \ref{isot_error}). Similar to \textsc{Big-Means}, its primary limitation is the accumulation of small errors across all the datasets and the numbers of clusters, (see, e.g., Tables \ref{index_mfcc} and \ref{index_skin}). Notably, \textsc{Big-Clust} finds the best known solution in only four cases, occurring for the following datasets: Shuttle Control ($k=2$), MiniBooNE Particle Identification ($k=2$), Range Queries Aggregates ($k=20$) and Covertype ($k=4$). On the other hand, \textsc{Big-Clust} is the fastest incremental algorithm across all numbers of clusters, as it achieves the highest number of runs with a CPU time less than or equal to 5 seconds (see Figures \ref{pienet_cpu} and \ref{isot_cpu}).

The incremental algorithm \textsc{DC-Clust} produces highly accurate results (see Figures \ref{pienet_error} and \ref{isot_error}). For instance, in the small $k$-clustering problems, \textsc{DC-Clust} achieves a relative error below 0.05\% in 59 cases (out of 72), whereas for the other five methods this number ranges from 0 to 56. Additionally, \textsc{DC-Clust} never produces a relative error exceeding 10\% with a small number of clusters, whereas the other methods show such high error rates in 3 to 26 cases. However, the CPU times of \textsc{DC-Clust} are significantly longer compared to the other methods (see Figures \ref{pienet_cpu} and \ref{isot_cpu}). For example, in the Covertype dataset, the total run time $t_\text{total}$ for \textsc{DC-Clust} was more than \emph{six hours} and 30 \emph{minutes}, whereas for the other five methods, $t_\text{total}$ ranged from 6.58 to 281.20 seconds. Moreover, within the 20-hour time limit, \textsc{DC-Clust} fails to find a solution for the Gisette dataset when $k>11$ and for the Music Analysis dataset when $k>15$.

Based on the results, \textsc{Clust-Splitter} and \textsc{LMBM-Clust} demonstrate the best overall performance, as the other four methods exhibit weaknesses in either computational time or accuracy. Both \textsc{Clust-Splitter} and \textsc{LMBM-Clust} frequently find the best known solution (72 and 68 times, respectively, out of a maximum of 144 cases) while maintaining relatively fast computation times, making them efficient in terms of both accuracy and execution time. In the small $k$-clustering problems, \textsc{LMBM-Clust} is slightly more accurate than \textsc{Clust-Splitter}, achieving 56 cases with a relative error below 0.05\%, compared to 53 cases with \textsc{Clust-Splitter} (see Figure \ref{pienet_error}). However, \textsc{Clust-Splitter} is slightly faster, with 35 cases having a CPU time below 5 seconds, compared to 33 cases for \textsc{LMBM-Clust} (see Figure \ref{pienet_cpu}). In the large $k$-clustering problems, \textsc{Clust-Splitter} outperforms \textsc{LMBM-Clust} in both accuracy and execution time: it achieves 22 cases with a relative error below 0.05\%, compared to 19 for \textsc{LMBM-Clust} (see Figure \ref{isot_error}), and has 2 cases with a CPU time below 1 second, while \textsc{LMBM-Clust} has none (see Figure \ref{isot_cpu}). Nevertheless, the differences between \textsc{Clust-Splitter} and \textsc{LMBM-Clust} remain relatively small.

\bigskip
\noindent \textbf{Detailed results}
\medskip

\noindent Now we move on to a more detailed analysis of the results, examining the performance of each method across different datasets and the numbers of clusters. When we compare the average relative errors $E_\text{aver}$ among all algorithms that successfully computed 25 clusters within the 20-hour time limit for each dataset, the following observations can be made. The \textsc{Clust-Splitter} algorithm achieves the lowest $E_\text{aver}$ in four datasets, \textsc{LMBM-Clust} in five datasets, \textsc{DC-Clust} in eight datasets, and \textsc{Big-Clust} in one dataset. Notably, \textsc{Big-Means} and \textsc{MiniBatchKMeans} do not achieve the lowest average error in any dataset. This highlights that all the best $E_\text{aver}$ values are obtained using the incremental algorithms.

Furthermore, if \textsc{DC-Clust} is excluded from the analysis due to its inability to compute 25 clusters in two datasets (Gisette and Music Analysis) and its significantly slower performance compared to other methods, the results change as follows: \textsc{Clust-Splitter} achieves the lowest $E_\text{aver}$ in nine datasets, \textsc{LMBM-Clust} in six datasets, \textsc{Big-Clust} in one dataset, and \textsc{Big-Means} in two datasets. Thus, we can conclude that \textsc{Clust-Splitter} and \textsc{LMBM-Clust} were the most accurate methods also in this type of analysis, with \textsc{Clust-Splitter} demonstrating slightly better accuracy.

For the small numbers of clusters ($k=2,3,4,5$), the following observations can be made regarding the number of times each method finds the best known solution (with a maximum possible count of 72 per method): \textsc{Clust-Splitter} achieves this outcome 51 times, \textsc{LMBM-Clust} 55 times, \textsc{DC-Clust} 58 times, \textsc{Big-Clust} 3 times, \textsc{Big-Means} 3 times, and \textsc{MiniBatchKMeans} does not find the best known solution in any instance. These findings indicate that, even with a small number of clusters, the incremental algorithms consistently outperform the other methods in minimizing the clustering error.

Since \textsc{Clust-Splitter} and \textsc{LMBM-Clust} appear to be the most accurate methods, we conduct a more detailed comparison of their performance. When examining the average relative error $E_\text{aver}$, \textsc{Clust-Splitter} performs better in 10 datasets, while \textsc{LMBM-Clust} outperforms it in 8 datasets. In terms of the total CPU time $t_\text{total}$, \textsc{Clust-Splitter} is faster in 3 datasets, whereas \textsc{LMBM-Clust} is faster in 15 datasets. However, when considering only the small $k$-clustering problems ($k=2,3,4,5$), \textsc{Clust-Splitter} has a lower computation time $t_k$ in 37 cases, while \textsc{LMBM-Clust} is faster in 35 cases. These results suggest that \textsc{Clust-Splitter} achieves a slightly higher average accuracy and performs marginally faster for the small numbers of clusters. However, when computing up to 25 clusters, \textsc{LMBM-Clust} demonstrates better efficiency in terms of computational time. Consequently, if a case involves a large number of clusters, \textsc{LMBM-Clust} is the preferable choice. Conversely, for problems with a smaller number of clusters, \textsc{Clust-Splitter} offers more accurate results with slightly faster computational times. The effectiveness of \textsc{Clust-Splitter} is further supported by its more intuitive selection of starting points compared to \textsc{LMBM-Clust}.

\bigskip
\noindent \textbf{Results with validity indices}
\medskip

\begin{figure}[ht!]
    \centering
    \includegraphics[width=\textwidth]{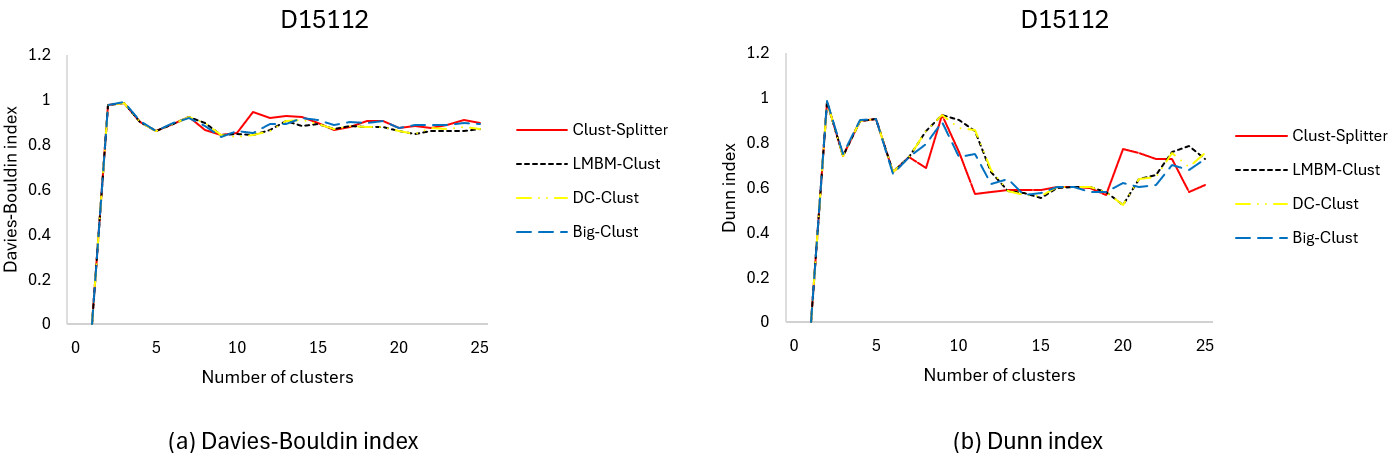}
    \caption{D15112: Davies-Bouldin and Dunn validity indices vs. number of clusters.}
    \label{index_d15112}
\end{figure}

\begin{figure}[ht!]
    \centering
    \includegraphics[width=\textwidth]{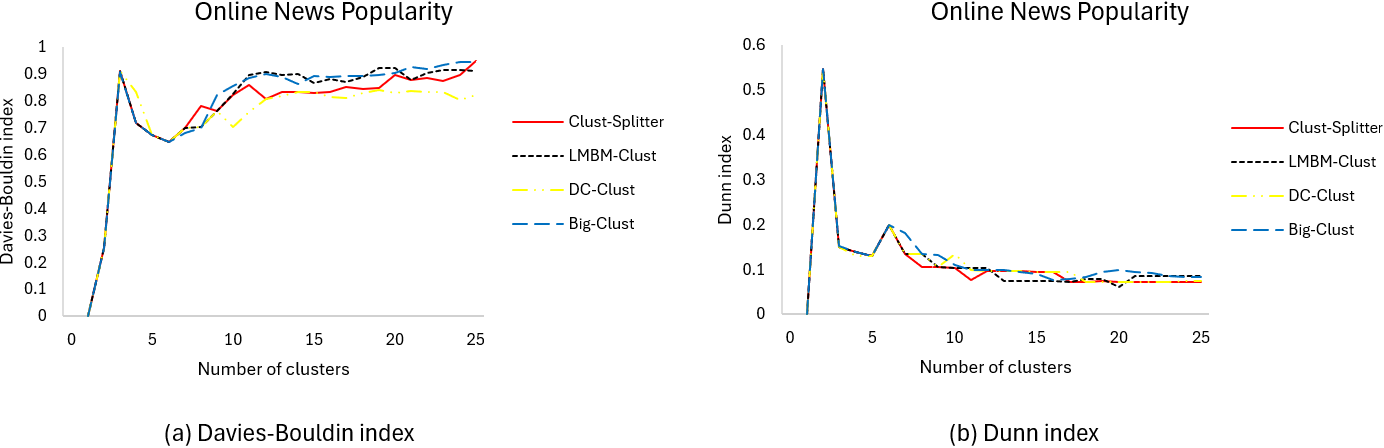}
    \caption{Online News Popularity: Davies-Bouldin and Dunn validity indices vs. number of clusters.}
    \label{index_onlinenews}
\end{figure}

\noindent The DBI measures how well-separated clusters are, with lower values indicating better separation. In nearly all DBI visualizations, the four incremental algorithms produce closely related values (see, e.g., Figures \ref{index_eegeye}$-$\ref{index_kegg}), though differences become more apparent as $k$ increases (see, e.g., Figures \ref{index_shuttle} and \ref{index_mfcc}). In 12 out of 18 datasets, \textsc{Clust-Splitter} produces DBI values identical to other methods for the small numbers of clusters ($k \leq 5$), with exceptions including ISOLET, Gas Sensor Array Drift, Shuttle Control, Sensorless Drive Diagnosis, MiniBooNE Particle Identification, and Protein Homology. These differences often arise from higher relative error values in \textsc{Clust-Splitter} (and in \textsc{Big-Means} and \textsc{MiniBatchKMeans}), suggesting that in some cases using partial data for clustering is not as effective as utilizing the full dataset. Furthermore, \textsc{Clust-Splitter} sometimes fails to find optimal solutions, resulting in higher DBI values (e.g., ISOLET at $k=3$, relative error 0.54). However, in other cases, it still identifies the best solution despite increased DBI values (e.g., Gas Sensor Array Drift at $k=4$). Importantly, even if \textsc{Clust-Splitter} produces a higher relative error for a specific $k$ value, this does not mean it cannot find an optimal solution at other $k$ values, including larger ones, for the same dataset. For instance, in the Protein Homology dataset, while the error is 1.82 at $k=2$, \textsc{Clust-Splitter} finds the best known solution at $k=3$ and $k=4$. Overall, the DBI values of \textsc{Clust-Splitter} align closely with the other methods especially for small values of $k$, reinforcing its competitiveness in identifying well-separated clusters. The DBI also suggests that optimal number of clusters is generally small, as seen for example in EEG Eye State ($k=4$) and MFCC Speech Emotion Recognition ($k=3$), where the lowest DBI values occur. 

\begin{figure}[ht!]
    \centering
    \includegraphics[width=\textwidth]{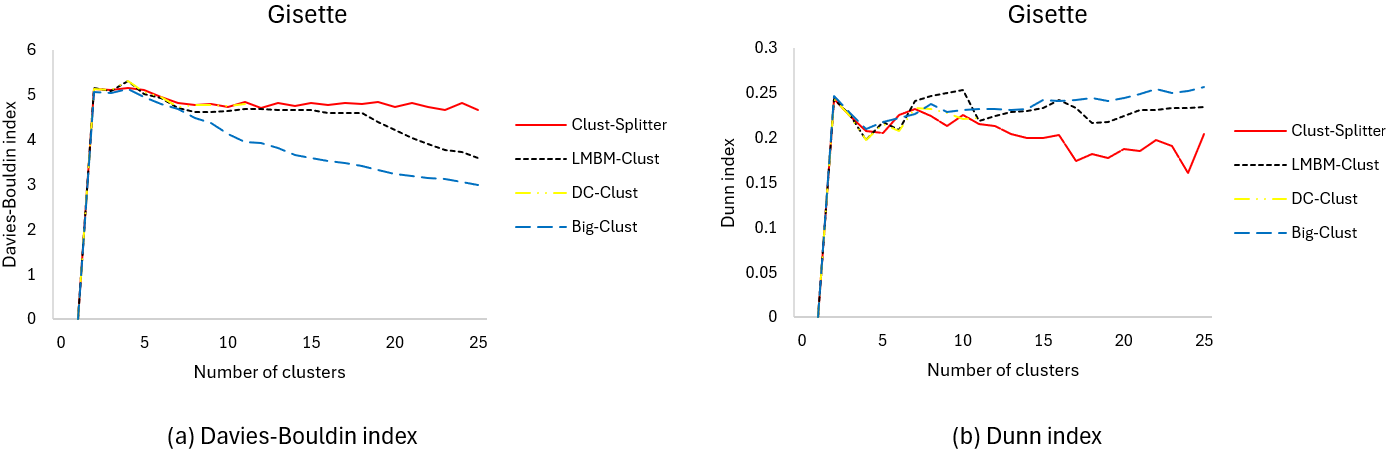}
    \caption{Gisette: Davies-Bouldin and Dunn validity indices vs. number of clusters.}
    \label{index_gisette}
\end{figure}

The DI, which evaluates cluster compactness and separation (with higher values indicating better clustering quality), exhibits a similar trend to the DBI. For small values of $k$, the incremental algorithms produce nearly identical DI values, confirming that \textsc{Clust-Splitter} effectively identifies compact, well-separated clusters. However, its performance declines as $k$ increases (see, e.g., Figures \ref{index_gisette} and \ref{index_mfcc}). In addition, DI visualizations indicate that the optimal number of clusters is small. For instance, Online News Popularity and KEGG Metabolic datasets favor $k=2$, while MFCC Speech Emotion Recognition dataset prefers $k=3$. Notably, \textsc{Clust-Splitter} determines the same optimal number of clusters as other methods in 14 out of 18 datasets, with exceptions including Gas Sensor Array Drift, Shuttle Control, Sensorless Drive Diagnosis, and Protein Homology.

\subsection{External validation and simulation study}

In addition to the internal validation indices DBI and DI, we assess the performance of \textsc{Clust-Splitter} using two external validity measures: the proportion of objects that are correctly grouped together against the true clusters and the \emph{adjusted Rand index} (ARI) \cite{HubAra:1985}. Furthermore, we evaluate how accurately the method groups data points according to their true clusters. This approach follows the methodology outlined in \cite{KarBagTah:2018}.

As mentioned, the well-known ARI is used to compare the clustering algorithms. Suppose that $U=\{U^1,U^2,\ldots,U^r\}$ represent the true partition of data points and $V=\{V^1,V^2,\ldots,V^s\}$ denote a clustering result. The ARI for $V$ is computed as
\begin{align*}
\text{ARI} = \frac{\sum_{ij} \binom{n_{ij}}{2} - \left[ \sum_i \binom{a_i}{2} \sum_j \binom{b_j}{2} \right] \big/ \binom{n}{2}}
{\frac{1}{2} \left[ \sum_i \binom{a_i}{2} + \sum_j \binom{b_j}{2} \right] - \left[ \sum_i \binom{a_i}{2} \sum_j \binom{b_j}{2} \right] \big/ \binom{n}{2}},
\end{align*}
where the notation $\binom{n_{ij}}{2}$ denotes the binomial coefficient '$n_{ij}$ choose $2$', $n$ is the total number of data points in the dataset, and $n_{ij}$, $a_i$, and $b_j$ are values derived from the contingency table given in Table \ref{table_arimaaritelma}. In this table, each entry $n_{ij}$ represents the number of data points shared between clusters $U_i$ and $V_j$. The ARI score ranges from $-1$ to $1$, where 1 indicates a perfect clustering match, 0 corresponds to random clustering, and negative values suggest a clustering result worse than random assignment.

\begin{table}[ht!]
\centering
\caption{Contingency table for comparing two partitions.} \vspace{0.25cm}
\begin{tabular}{l c c c c c}
\toprule
& $V^1$ & $V^2$ & $\ldots$ & $V^s$ & Sums \\
\midrule
$U^1$ & $n_{11}$ & $n_{12}$ & $\ldots$ & $n_{1s}$ & $a_1$ \\
\midrule
$U^2$ & $n_{21}$ & $n_{22}$ & $\ldots$ & $n_{2s}$ & $a_2$ \\
\midrule
 & $\vdots$ & $\vdots$ & $\ddots$ & $\vdots$ & $\vdots$ \\
\midrule
$U^r$ & $n_{r1}$ & $n_{r2}$ & $\ldots$ & $n_{rs}$ & $a_r$ \\
\midrule
Sums & $b_1$ & $b_2$ & $\ldots$ & $b_s$ &  \\
\bottomrule
\end{tabular}
\label{table_arimaaritelma}
\end{table}

\subsubsection{Validation with real-world datasets}

First, we apply \textsc{Clust-Splitter}, \textsc{LMBM-Clust}, \textsc{DC-Clust}, \textsc{Big-Clust}, \textsc{Big-Means}, and \textsc{MiniBatchKMeans} to three real-world datasets with known true cluster labels: Iris, Soybean, and Arcane (training set only). These datasets are publicly available from \cite{uci}. The Iris dataset consists of 150 data points with four attributes and is classified into three clusters, each containing 50 data points. The Soybean dataset comprises 47 data points with 35 attributes and is divided into four clusters, where the first three clusters contain 10 data points each, while the fourth cluster has 17 data points. The Arcane dataset includes 100 data points with 10,000 attributes and is split into two clusters, with 44 data points in the first cluster and 56 in the second.

In our experiments, the number of clusters was set to match the known number of clusters in each original dataset. The algorithms were applied to the datasets without access to the known cluster labels. The performance of each method was evaluated based on accuracy, defined as the proportion of data points correctly grouped together against to the true clusters. In addition, we computed the ARI to further assess clustering quality. The results obtained for the different algorithms are presented in Tables \ref{ari_iris}\,--\,\ref{ari_arcane}.

\begin{table}[ht!]
\centering
\caption{Clustering results with Iris.} \vspace{0.25cm}
\resizebox{\textwidth}{!}{ 
\begin{tabular}{c ccc ccc ccc ccc ccc ccc}
\toprule
\text{True cluster} 
& \multicolumn{3}{c}{\textsc{Clust-Splitter}} 
& \multicolumn{3}{c}{\textsc{LMBM-Clust}} 
& \multicolumn{3}{c}{\textsc{DC-Clust}}
& \multicolumn{3}{c}{\textsc{Big-Clust}}
& \multicolumn{3}{c}{\textsc{Big-Means}}
& \multicolumn{3}{c}{\textsc{MiniBatchKMeans}} \\
\cmidrule(lr){2-4} \cmidrule(lr){5-7} \cmidrule(lr){8-10} \cmidrule(lr){11-13} \cmidrule(lr){14-16} \cmidrule(lr){17-19}
& 1 & 2 & 3  
& 1 & 2 & 3  
& 1 & 2 & 3  
& 1 & 2 & 3  
& 1 & 2 & 3  
& 1 & 2 & 3  \\
\midrule
1 & 50	&	0	&	0	&	47	&	0	&	3	&	47	&	0	&	3	&	48	&	0	&	2	&	48	&	0	&	2	&	47	&	0	&	3 \\
2	&	0	&	50	&	0	&	0	&	50	&	0	&	0	&	50	&	0	&	0	&	50	&	0	&	0	&	50	&	0	&	0	&	50	&	0	\\
3	&	15	&	0	&	35	&	14	&	0	&	36	&	14	&	0	&	36	&	14	&	0	&	36	&	14	&	0	&	36	&	14	&	0	&	36	\\
\midrule
Accuracy (\%)	&	90.0 &&&
88.7 &&&
88.7 &&&
89.3 &&&
89.3 &&&
88.7 \\
ARI	&	0.7455 &&&
0.7163 &&&
0.7163 &&&
0.7302 &&&
0.7302 &&&
0.7163 \\
\bottomrule
\end{tabular}
\label{ari_iris}
} 
\end{table}

\begin{table}[ht!]
\centering
\caption{Clustering results with Soybean.} \vspace{0.25cm}
\resizebox{\textwidth}{!}{ 
\begin{tabular}{c cccc cccc cccc cccc cccc cccc}
\toprule
\text{True cluster} 
& \multicolumn{4}{c}{\textsc{Clust-Splitter}} 
& \multicolumn{4}{c}{\textsc{LMBM-Clust}} 
& \multicolumn{4}{c}{\textsc{DC-Clust}}
& \multicolumn{4}{c}{\textsc{Big-Clust}}
& \multicolumn{4}{c}{\textsc{Big-Means}}
& \multicolumn{4}{c}{\textsc{MiniBatchKMeans}} \\
\cmidrule(lr){2-5} \cmidrule(lr){6-9} \cmidrule(lr){10-13} \cmidrule(lr){14-17} \cmidrule(lr){18-21} \cmidrule(lr){22-25}
& 1 & 2 & 3 & 4  
& 1 & 2 & 3 & 4  
& 1 & 2 & 3 & 4  
& 1 & 2 & 3 & 4  
& 1 & 2 & 3 & 4  
& 1 & 2 & 3 & 4  \\
\midrule
1 & 5	&	0	&	0	&	5	&	5	&	0	&	0	&	5	&	5	&	0	&	0	&	5	&	5	&	0	&	0	&	5	&	5	&	0	&	0	&	5	&	9	&	0	&	0	&	1 \\
2 & 0	&	10	&	0	&	0	&	0	&	10	&	0	&	0	&	0	&	10	&	0	&	0	&	0	&	10	&	0	&	0	&	0	&	10	&	0	&	0	&	0	&	10	&	0	&	0 \\
3 & 0	&	0	&	10	&	0	&	0	&	0	&	10 &	0	&	0	&	0	&	10	&	0	&	0	&	0	&	10	&	0	&	0	&	0	&	10	&	0	&	0	&	0	&	10	&	0 \\
4 & 5	&	0	&	0	&	12	&	0	&	0	&	0 &	17	&	7	&	0	&	0	&	10	&	6	&	0	&	0	&	11	&	8	&	0	&	0	&	9	&	5	&	0	&	0	&	12 \\
\midrule
Accuracy (\%)	&	78.7 &&&&
89.4 &&&&
74.5 &&&&
76.6 &&&&
72.3 &&&&
87.2 \\
ARI &  
0.5814 &&&&
0.7477 &&&&
0.5513 &&&&
0.5634 &&&&
0.5452 &&&&
0.6851 \\
\bottomrule
\end{tabular}
\label{ari_soybean}
} 
\end{table}

\begin{table}[ht!]
\centering
\caption{Clustering results with Arcane (training set only).} \vspace{0.25cm}
\resizebox{\textwidth}{!}{ 
\begin{tabular}{c cc cc cc cc cc cc}
\toprule
\text{True cluster} 
& \multicolumn{2}{c}{\textsc{Clust-Splitter}} 
& \multicolumn{2}{c}{\textsc{LMBM-Clust}} 
& \multicolumn{2}{c}{\textsc{DC-Clust}}
& \multicolumn{2}{c}{\textsc{Big-Clust}}
& \multicolumn{2}{c}{\textsc{Big-Means}}
& \multicolumn{2}{c}{\textsc{MiniBatchKMeans}} \\
\cmidrule(lr){2-3} \cmidrule(lr){4-5} \cmidrule(lr){6-7} \cmidrule(lr){8-9} \cmidrule(lr){10-11} \cmidrule(lr){12-13}
& 1 & 2  
& 1 & 2  
& 1 & 2  
& 1 & 2  
& 1 & 2  
& 1 & 2  \\
\midrule
1 & 17	&	27	&	17	&	27	&	17	&	27	&	17	&	27	&	17	&	27	&	17	&	27 \\
2 & 21	&	35	&	22	&	34	&	21	&	35	&	22	&	34	&	22	&	34	&	21	&	35 \\
\midrule
Accuracy (\%)	&	52.0 &&
51.0 &&
52.0 &&
51.0 &&
51.0 &&
52.0 \\
ARI & 
-0.0087 &&
-0.0099 &&
-0.0087 &&
-0.0099 &&
-0.0099 &&
-0.0087 \\
\bottomrule
\end{tabular}
\label{ari_arcane}
} 
\end{table}

The pairwise comparison of the clustering algorithms demonstrates that the \textsc{Clust-Splitter} method mainly achieves improved accuracy while it also yields in the most cases the highest ARI values. In the Iris dataset, \textsc{Clust-Splitter} attains the highest accuracy at 90.0\%. For the Soybean dataset, it ranks third in accuracy, achieving 78.7\%. In the Arcane dataset, \textsc{Clust-Splitter} achieves the highest accuracy at 52.0\%. However, in the Arcane dataset, the negative ARI values indicate that clustering results are worse than random clustering across all the methods.

Overall, all the methods exhibit relatively similar performance across the datasets. The difference in accuracy within the Iris dataset is 1.3\%, corresponding to the reassignment of three data points. In the Soybean dataset, the accuracy difference between the best and worst-performing methods is 17.1\%, which translates to eight data points. In the Arcane dataset, the difference is 1.0\%, meaning that one data point was assigned differently. In summary, \textsc{Clust-Splitter} has demonstrated strong and accurate performance in these tests. Nevertheless, the differences between the algorithms remain relatively small.

\subsubsection{Study with simulated data}

The effectiveness of the \textsc{Clust-Splitter} method is further assessed using artificially generated datasets. These datasets were created in a two-dimensional space, incorporating varying proportions of outliers. Each dataset consists of three clusters (denoted as $A$, $B$, and $C$), with each cluster containing 120 data points. The coordinates of data points for the cluster $A$ are sampled independently from the normal distributions $N(\mu_A^x, \sigma_A)$ and $N(\mu_A^y, \sigma_A)$, where $\mu_A^x$ and $\mu_A^y$ represent the mean values, and $\sigma_A$ denotes the standard deviation. Similarly, the coordinates of data points for the clusters $B$ and $C$ are drawn from the corresponding normal distributions $N(\mu_B^x, \sigma_B)$, $N(\mu_B^y, \sigma_B)$, $N(\mu_C^x, \sigma_C)$, and $N(\mu_C^y, \sigma_C)$. In the cluster $C$, a designated proportion of data points, referred to as outliers, exhibit a greater standard deviation $\sigma_C^O$ compared to the standard deviation $\sigma_C$ of the remaining data points in the cluster. For instance, if 20\% of the data points in the cluster $C$ are outliers, then the coordinates of 24 data points are sampled from the normal distributions with $\sigma_C^O$, while the remaining 96 points follow the standard deviation $\sigma_C$. The parameters used to generate these simulated datasets are detailed in Table \ref{table_parameters}. Notably, the same parameter values as in \cite{KarBagTah:2018} were used. To ensure robustness, datasets with varying proportions of outliers were generated, and for each proportion, ten different datasets were created. The results, presented in Table \ref{table_generoitu}, represent the average performance across these ten datasets.  

\begin{table}[ht!]
\centering
\caption{Parameters for generating data points in artificial data.} \vspace{0.25cm}

\begin{tabular}{l c c c}
\toprule
& \text{Cluster $A$} & \text{Cluster $B$} & \text{Cluster $C$} \\
\midrule
Mean $\mu^x$ & 0   & 6   & 6   \\
Mean $\mu^y$ & 0   & $-1$  & 2   \\
Standard deviation $\sigma$ & 1.5 & 0.5 & 0.5 \\
Outliers $\sigma_C^O$ & $-$  & $-$  & 2   \\
\bottomrule
\end{tabular}
\label{table_parameters}
\end{table}

\begin{table}[ht!]
\centering
\caption{ARI values by different clustering algorithms in artificial data.} \vspace{0.25cm}
\resizebox{\textwidth}{!}{ 
\begin{tabular}{c c c c c c c}
\toprule
\text{Outliers (\%)} 
& \textsc{Clust-Splitter}
& \textsc{LMBM-Clust}
& \textsc{DC-Clust}
& \textsc{Big-Clust}
& \textsc{Big-Means}
& \textsc{MiniBatchKMeans} \\
\midrule
0 \%	&	0.9784	&	0.9776	&	0.9784	&	0.9705	&	0.9239	&	0.9238	\\
10 \%	&	0.9407	&	0.9415	&	0.9398	&	0.9407	&	0.8907	&	0.8890	\\
20 \%	&	0.9109	&	0.9141	&	0.9125	&	0.9108	&	0.8606	&	0.9125	\\
30 \%	&	0.8882	&	0.8890	&	0.8890	&	0.8874	&	0.8420	&	0.8416	\\
40 \%	&	0.8712	&	0.8706	&	0.8729	&	0.8750	&	0.8239	&	0.6510	\\
50 \%	&	0.8272	&	0.8250	&	0.8265	&	0.8279	&	0.7906	&	0.8248	\\
\bottomrule
\end{tabular}
\label{table_generoitu}
} 
\end{table}

Table \ref{table_generoitu} shows that the ARI values for the different clustering algorithms are very close to each other. In particular, the four incremental algorithms $-$ \textsc{Clust-Splitter}, \textsc{LMBM-Clust}, \textsc{DC-Clust}, and \textsc{Big-Clust} $-$ exhibit highly similar ARI values. In contrast, the two other algorithms, \textsc{Big-Means} and \textsc{MiniBatchKMeans}, also yield comparable results to each other but with lower ARI values than obtained with the incremental algorithms, indicating weaker clustering performance. Overall, the differences in ARI values among the incremental algorithms were minimal, demonstrating that \textsc{Clust-Splitter} performed well in the tests, suggesting that it is capable of handling outliers effectively.

\section{Conclusions} \label{kappale6}

In this paper, we introduce a new clustering method, \textsc{Clust-Splitter}, for solving minimum sum-of-squares clustering problems, particularly in large-scale datasets. \textsc{Clust-Splitter} is an incremental algorithm, meaning that, in addition to solving the $k_{max}$-clustering problem for a given number of clusters $k_{max}$, it also solves all intermediate $k$-clustering problems for $k=1,\ldots,k_{max}-1$. The method consists of three main components: a novel approach for selecting starting points based on cluster splitting with the help of auxiliary clustering problems, an incremental clustering algorithm and the limited memory bundle method, which is employed at each iteration to solve both the clustering and auxiliary clustering problems.

The proposed \textsc{Clust-Splitter} method was tested on 18 large-scale real-world datasets, with the number of data points ranging from tens of thousands to up to hundreds of thousands. It was compared against state-of-the-art clustering methods, including \textsc{LMBM-Clust}, \textsc{DC-Clust}, \textsc{Big-Clust}, \textsc{Big-Means}, and \textsc{MiniBatchKMeans}, using various evaluation metrics such as sum-of-squares errors, two cluster validity indices, and CPU time. The results show that \textsc{Clust-Splitter} is highly efficient and effective for large-scale clustering problems.

The results presented in this paper demonstrate that \textsc{Clust-Splitter} and \textsc{LMBM-Clust} are the most effective methods, as they achieve excellent accuracy while maintaining fast computation times. In contrast, \textsc{MiniBatchKMeans} is the fastest method but performs poorly in terms of accuracy. On the other hand, \textsc{DC-Clust} produces highly accurate solutions but requires significantly longer computation times. \textsc{Big-Clust} and \textsc{Big-Means} solve clustering problems relatively quickly and come close to the best known solution, but they rarely achieve it. Since \textsc{Clust-Splitter} and \textsc{LMBM-Clust} frequently reach the best known solution within a reasonable time, they can be considered the best-performing methods in this study.

Additionally, while \textsc{LMBM-Clust} is the preferred choice for clustering more than ten clusters, \textsc{Clust-Splitter} is better suited for problems with small number of clusters. It efficiently identifies compact and well-separated clusters with low CPU time while often achieving or closely approximating the best known clustering function value. This advantage is further reinforced by the more intuitive selection of starting points used by \textsc{Clust-Splitter} compared to \textsc{LMBM-Clust}, making it a practical choice for clustering tasks. Overall, these findings indicate that \textsc{Clust-Splitter} is a strong competitor alongside \textsc{LMBM-Clust}, particularly for real-time clustering in large-scale datasets.

\section*{Acknowledgement}
The work was financially supported by the Research Council of Finland (projects no. \#345804 and \#345805 led by Prof.\@ Tapio Pahikkala and Prof.\@ Antti Airola, respectively), and Jenny and Antti Wihuri Foundation.

\bibliographystyle{abbrv}
\bibliography{references}

\newpage

\section*{Appendix}

\begin{table}[ht!]
\centering
\caption{Summary of the results with ISOLET ($\times 10^5$). Dimensions: $m$ = 7797, $n$ = 617.} \vspace{0.25cm}
\resizebox{\textwidth}{!}{ 
\begin{tabular}{
    c c 
    c c
    c c
    c c
    c c
    c c
    c c
}
\toprule
$k$ & $f_\text{best}$ 
& \multicolumn{2}{c}{\textsc{Clust-Splitter}} 
& \multicolumn{2}{c}{\textsc{LMBM-Clust}} 
& \multicolumn{2}{c}{\textsc{DC-Clust}}
& \multicolumn{2}{c}{\textsc{Big-Clust}}
& \multicolumn{2}{c}{\textsc{Big-Means}}
& \multicolumn{2}{c}{\textsc{MiniBatchKMeans}} \\
\cmidrule(lr){3-4} \cmidrule(lr){5-6} \cmidrule(lr){7-8} \cmidrule(lr){9-10} \cmidrule(lr){11-12} \cmidrule(lr){13-14}
& & \multicolumn{1}{c}{$E_k$} & \multicolumn{1}{c}{$t_k$} 
& \multicolumn{1}{c}{$E_k$} & \multicolumn{1}{c}{$t_k$} 
& \multicolumn{1}{c}{$E_k$} & \multicolumn{1}{c}{$t_k$} 
& \multicolumn{1}{c}{$E_k$} & \multicolumn{1}{c}{$t_k$} 
& \multicolumn{1}{c}{$E_k$} & \multicolumn{1}{c}{$t_k$} 
& \multicolumn{1}{c}{$E_k$} & \multicolumn{1}{c}{$t_k$} \\
\midrule
2	&	7.20988	&	0.00	&	6.02	&	0.00	&	4.11	&	0.00	&	66.63	&	0.26	&	4.60	&	0.13	&	4.74	&	0.27	&	0.13	\\
3	&	6.77921	&	0.54	&	11.20	&	0.00	&	8.70	&	0.00	&	131.80	&	0.37	&	8.94	&	0.40	&	4.63	&	0.79	&	0.12	\\
4	&	6.41487*	&	0.00	&	13.14	&	0.60	&	12.06	&	0.60	&	202.30	&	0.45	&	12.70	&	0.35	&	4.67	&	1.28	&	0.14	\\
5	&	6.12976	&	0.00	&	16.53	&	0.00	&	17.03	&	1.01	&	271.64	&	0.49	&	18.29	&	0.50	&	4.59	&	1.68	&	0.16	\\
10	&	5.28577	&	0.06	&	31.17	&	2.72	&	35.84	&	1.18	&	649.78	&	2.11	&	33.48	&	1.13	&	4.69	&	2.64	&	0.17	\\
15	&	4.86785	&	0.89	&	44.34	&	0.10	&	58.47	&	0.00	&	1068.70	&	2.57	&	51.17	&	1.28	&	4.71	&	2.30	&	0.18	\\
20	&	4.60260	&	0.46	&	63.48	&	0.02	&	81.55	&	0.00	&	1606.58	&	2.56	&	73.25	&	1.69	&	4.70	&	2.90	&	0.20	\\
25	&	4.43890	&	0.30	&	88.52	&	0.58	&	106.28	&	0.27	&	2196.53	&	2.59	&	98.13	&	0.89	&	4.70	&	2.49	&	0.20	\\

\midrule
$E_\text{aver}$	&		&	0.28	&		&	0.50	&		&	0.38	&		&	1.42	&		&	0.80	&		&	1.80	&		\\
$t_\text{init}$	&		&		&	6.52	&		&	6.22	&		&	6.23	&		&	6.28	&		&	1.22	&		&	1.17	\\
$t_\text{total}$	&		&		&	95.03	&		&	112.50	&		&	2202.77	&		&	104.40	&		&	38.65	&		&	2.47	\\

\bottomrule
\end{tabular}
\label{taulukko_isolet}
} 
\end{table}

\begin{table}[ht!]
\centering
\caption{Summary of the results with Gisette ($\times 10^{12}$). Dimensions: $m$ = 13,500, $n$ = 5000.} \vspace{0.25cm}
\resizebox{\textwidth}{!}{ 
\begin{tabular}{
    c c 
    c c
    c c
    c c
    c c
    c c
    c c
}
\toprule
$k$ & $f_\text{best}$ 
& \multicolumn{2}{c}{\textsc{Clust-Splitter}} 
& \multicolumn{2}{c}{\textsc{LMBM-Clust}} 
& \multicolumn{2}{c}{\textsc{DC-Clust}}
& \multicolumn{2}{c}{\textsc{Big-Clust}}
& \multicolumn{2}{c}{\textsc{Big-Means}}
& \multicolumn{2}{c}{\textsc{MiniBatchKMeans}} \\
\cmidrule(lr){3-4} \cmidrule(lr){5-6} \cmidrule(lr){7-8} \cmidrule(lr){9-10} \cmidrule(lr){11-12} \cmidrule(lr){13-14}
& & \multicolumn{1}{c}{$E_k$} & \multicolumn{1}{c}{$t_k$} 
& \multicolumn{1}{c}{$E_k$} & \multicolumn{1}{c}{$t_k$} 
& \multicolumn{1}{c}{$E_k$} & \multicolumn{1}{c}{$t_k$} 
& \multicolumn{1}{c}{$E_k$} & \multicolumn{1}{c}{$t_k$} 
& \multicolumn{1}{c}{$E_k$} & \multicolumn{1}{c}{$t_k$} 
& \multicolumn{1}{c}{$E_k$} & \multicolumn{1}{c}{$t_k$} \\
\midrule
2	&	4.19944	&	0.00	&	23.58	&	0.00	&	48.66	&	0.00	&	4401.03	&	0.17	&	27.63	&	0.01	&	27.18	&	0.30	&	0.95	\\
3	&	4.11596	&	0.00	&	182.97	&	0.00	&	103.64	&	0.00	&	9417.34	&	0.27	&	53.58	&	0.02	&	27.14	&	0.35	&	0.95	\\
4	&	4.06539	&	0.04	&	310.44	&	0.00	&	164.19	&	0.00	&	15,448.19	&	0.38	&	82.75	&	0.08	&	27.19	&	0.37	&	0.98	\\
5	&	4.02235*	&	0.00	&	404.20	&	0.02	&	236.16	&	0.02	&	21,447.47	&	0.49	&	110.61	&	0.09	&	26.65	&	0.38	&	1.21	\\
10	&	3.87672	&	0.02	&	940.64	&	0.20	&	644.52	&	0.03	&	57,633.72	&	1.70	&	290.43	&	0.16	&	27.40	&	0.61	&	1.29	\\
15	&	3.80098*	&	0.00	&	1603.30	&	0.14	&	1207.41	&	$-$	&	$-$	&	2.39	&	523.98	&	0.10	&	27.57	&	0.59	&	1.23	\\
20	&	3.74592*	&	0.00	&	2554.77	&	0.43	&	1842.69	&	$-$	&	$-$	&	3.01	&	847.30	&	0.13	&	27.81	&	0.60	&	1.58	\\
25	&	3.70152*	&	0.00	&	3634.23	&	1.51	&	2229.50	&	$-$	&	$-$	&	3.62	&	1285.57	&	0.20	&	27.68	&	0.66	&	1.48	\\

\midrule
$E_\text{aver}$	&		&	0.01	&		&	0.29	&		&	0.01	&		&	1.50	&		&	0.10	&		&	0.48	&		\\
$t_\text{init}$	&		&		&	64.30	&		&	61.89	&		&	63.55	&		&	61.86	&		&	21.12	&		&	17.90	\\
$t_\text{total}$	&		&		&	3698.53	&		&	2291.39	&		&	57,697.27	&		&	1347.43	&		&	239.75	&		&	27.57	\\

\bottomrule
\end{tabular}
\label{taulukko_gisette}
} 
\end{table}

\begin{table}[ht!]
\centering
\caption{Summary of the results with Gas Sensor Array Drift ($\times 10^{13}$). Dimensions: $m$ = 13,910, $n$ = 128.} \vspace{0.25cm}
\resizebox{\textwidth}{!}{ 
\begin{tabular}{
    c c 
    c c
    c c
    c c
    c c
    c c
    c c
}
\toprule
$k$ & $f_\text{best}$ 
& \multicolumn{2}{c}{\textsc{Clust-Splitter}} 
& \multicolumn{2}{c}{\textsc{LMBM-Clust}} 
& \multicolumn{2}{c}{\textsc{DC-Clust}}
& \multicolumn{2}{c}{\textsc{Big-Clust}}
& \multicolumn{2}{c}{\textsc{Big-Means}}
& \multicolumn{2}{c}{\textsc{MiniBatchKMeans}} \\
\cmidrule(lr){3-4} \cmidrule(lr){5-6} \cmidrule(lr){7-8} \cmidrule(lr){9-10} \cmidrule(lr){11-12} \cmidrule(lr){13-14}
& & \multicolumn{1}{c}{$E_k$} & \multicolumn{1}{c}{$t_k$} 
& \multicolumn{1}{c}{$E_k$} & \multicolumn{1}{c}{$t_k$} 
& \multicolumn{1}{c}{$E_k$} & \multicolumn{1}{c}{$t_k$} 
& \multicolumn{1}{c}{$E_k$} & \multicolumn{1}{c}{$t_k$} 
& \multicolumn{1}{c}{$E_k$} & \multicolumn{1}{c}{$t_k$} 
& \multicolumn{1}{c}{$E_k$} & \multicolumn{1}{c}{$t_k$} \\
\midrule
2	&	7.91186*	&	0.00	&	3.91	&	0.00	&	27.72	&	0.00	&	11.25	&	0.03	&	0.63	&	0.07	&	4.63	&	0.74	&	0.06	\\
3	&	5.02412	&	0.00	&	7.70	&	0.00	&	32.14	&	0.00	&	27.73	&	0.10	&	1.61	&	0.13	&	4.64	&	2.97	&	0.05	\\
4	&	3.97506*	&	0.00	&	11.20	&	4.61	&	32.94	&	4.63	&	50.75	&	4.75	&	2.22	&	1.49	&	4.64	&	3.64	&	0.05	\\
5	&	3.22394	&	0.00	&	12.08	&	0.00	&	34.53	&	0.10	&	75.73	&	0.12	&	2.89	&	4.60	&	4.23	&	13.12	&	0.04	\\
10	&	1.65230	&	0.18	&	23.23	&	0.18	&	48.50	&	0.18	&	245.08	&	5.85	&	5.87	&	3.52	&	4.65	&	26.89	&	0.06	\\
15	&	1.13801	&	2.62	&	45.89	&	0.36	&	56.98	&	0.36	&	500.20	&	0.58	&	8.94	&	4.20	&	4.24	&	18.89	&	0.09	\\
20	&	0.87916	&	1.99	&	68.27	&	2.79	&	63.45	&	0.61	&	771.25	&	3.07	&	12.89	&	5.37	&	4.54	&	17.02	&	0.08	\\
25	&	0.72211	&	2.65	&	88.25	&	4.31	&	70.48	&	0.65	&	1051.48	&	3.46	&	17.33	&	4.61	&	4.66	&	23.73	&	0.06	\\

\midrule
$E_\text{aver}$	&		&	0.93	&		&	1.53	&		&	0.82	&		&	2.24	&		&	3.00	&		&	13.38	&		\\
$t_\text{init}$	&		&		&	2.53	&		&	2.42	&		&	2.44	&		&	2.50	&		&	0.47	&		&	0.45	\\
$t_\text{total}$	&		&		&	90.78	&		&	72.91	&		&	1053.92	&		&	19.83	&		&	36.69	&		&	0.94	\\

\bottomrule
\end{tabular}
\label{taulukko_gas}
} 
\end{table}

\begin{figure}[ht!]
    \centering
    \includegraphics[width=\textwidth]{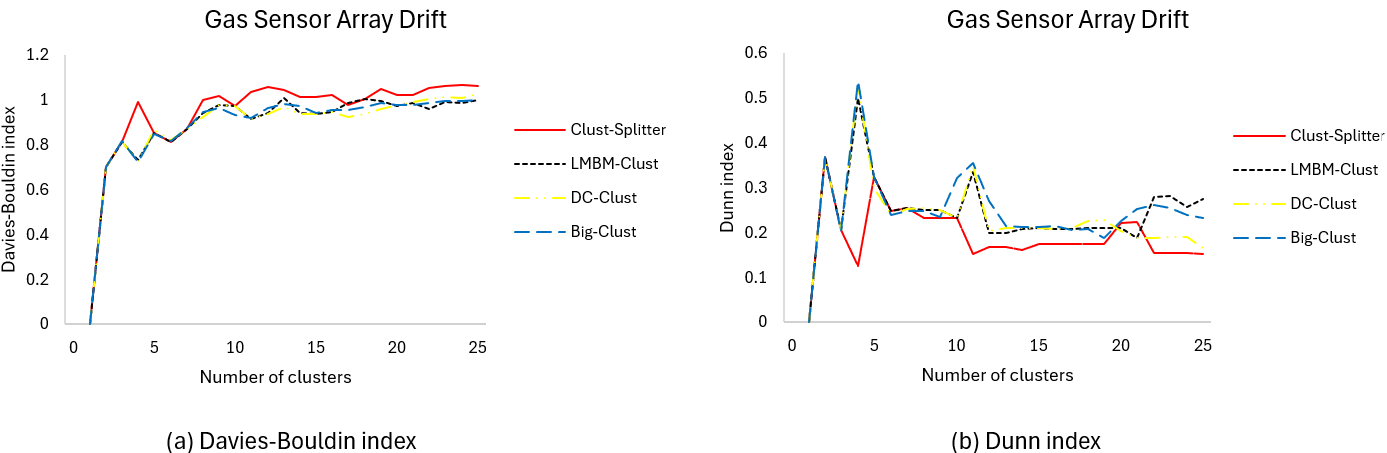}
    \caption{Gas Sensor Array Drift: Davies-Bouldin and Dunn validity indices vs. number of clusters.}
    \label{index_gas}
\end{figure}

\begin{table}[ht!]
\centering
\caption{Summary of the results with EEG Eye State ($\times 10^8$). Dimensions: $m$ = 14,980, $n$ = 14.} \vspace{0.25cm}
\resizebox{\textwidth}{!}{ 
\begin{tabular}{
    c c 
    c c
    c c
    c c
    c c
    c c
    c c
}
\toprule
$k$ & $f_\text{best}$ 
& \multicolumn{2}{c}{\textsc{Clust-Splitter}} 
& \multicolumn{2}{c}{\textsc{LMBM-Clust}} 
& \multicolumn{2}{c}{\textsc{DC-Clust}}
& \multicolumn{2}{c}{\textsc{Big-Clust}}
& \multicolumn{2}{c}{\textsc{Big-Means}}
& \multicolumn{2}{c}{\textsc{MiniBatchKMeans}} \\
\cmidrule(lr){3-4} \cmidrule(lr){5-6} \cmidrule(lr){7-8} \cmidrule(lr){9-10} \cmidrule(lr){11-12} \cmidrule(lr){13-14}
& & \multicolumn{1}{c}{$E_k$} & \multicolumn{1}{c}{$t_k$} 
& \multicolumn{1}{c}{$E_k$} & \multicolumn{1}{c}{$t_k$} 
& \multicolumn{1}{c}{$E_k$} & \multicolumn{1}{c}{$t_k$} 
& \multicolumn{1}{c}{$E_k$} & \multicolumn{1}{c}{$t_k$} 
& \multicolumn{1}{c}{$E_k$} & \multicolumn{1}{c}{$t_k$} 
& \multicolumn{1}{c}{$E_k$} & \multicolumn{1}{c}{$t_k$} \\
\midrule
2	&	7845.09934	&	4.25	&	0.89	&	4.25	&	0.05	&	4.25	&	0.13	&	4.25	&	0.09	&	62.02	&	3.25	&	98.14	&	0.03	\\
3	&	1833.88058	&	0.00	&	2.20	&	0.00	&	0.08	&	0.00	&	0.33	&	0.01	&	0.14	&	663.56	&	3.30	&	746.99	&	0.02	\\
4	&	2.23605	&	0.00	&	3.00	&	0.00	&	0.11	&	0.00	&	0.63	&	0.01	&	0.18	&	510,690.30	&	3.28	&	68,9042.19	&	0.03	\\
5	&	1.33858	&	0.00	&	3.19	&	0.00	&	0.17	&	0.00	&	1.33	&	0.01	&	0.25	&	882,321.25	&	3.39	&	1,154,863.91	&	0.03	\\
10	&	0.45306*	&	0.00	&	4.09	&	0.80	&	2.25	&	0.81	&	9.05	&	1.06	&	0.80	&	2,583,946.58	&	3.33	&	3,417,200.94	&	0.03	\\
15	&	0.34653	&	0.62	&	5.78	&	0.05	&	4.73	&	0.26	&	22.50	&	1.78	&	1.34	&	3,202,194.92	&	3.27	&	4,376,517.58	&	0.03	\\
20	&	0.28986	&	0.75	&	9.47	&	0.00	&	7.58	&	1.34	&	44.09	&	2.41	&	1.98	&	3,000,995.30	&	3.35	&	5,212,690.01	&	0.03	\\
25	&	0.25989	&	0.29	&	13.11	&	0.16	&	11.50	&	0.13	&	68.53	&	2.51	&	2.73	&	5,142,507.29	&	3.28	&	5,723,176.90	&	0.03	\\

\midrule
$E_\text{aver}$	&		&	0.74	&		&	0.66	&		&	0.85	&		&	1.51	&		&	1,915,422.65	&		&	2,571,792.08	&		\\
$t_\text{init}$	&		&		&	0.30	&		&	0.30	&		&	0.30	&		&	0.31	&		&	0.06	&		&	0.06	\\
$t_\text{total}$	&		&		&	13.41	&		&	11.80	&		&	68.83	&		&	3.05	&		&	26.52	&		&	0.28	\\

\bottomrule
\end{tabular}
\label{taulukko_eegeye}
} 
\end{table}

\begin{table}[ht!]
\centering
\caption{Summary of the results with D15112 ($\times 10^{11}$). Dimensions: $m$ = 15,112, $n$ = 2.} \vspace{0.25cm}
\resizebox{\textwidth}{!}{ 
\begin{tabular}{
    c c 
    c c
    c c
    c c
    c c
    c c
    c c
}
\toprule
$k$ & $f_\text{best}$ 
& \multicolumn{2}{c}{\textsc{Clust-Splitter}} 
& \multicolumn{2}{c}{\textsc{LMBM-Clust}} 
& \multicolumn{2}{c}{\textsc{DC-Clust}}
& \multicolumn{2}{c}{\textsc{Big-Clust}}
& \multicolumn{2}{c}{\textsc{Big-Means}}
& \multicolumn{2}{c}{\textsc{MiniBatchKMeans}} \\
\cmidrule(lr){3-4} \cmidrule(lr){5-6} \cmidrule(lr){7-8} \cmidrule(lr){9-10} \cmidrule(lr){11-12} \cmidrule(lr){13-14}
& & \multicolumn{1}{c}{$E_k$} & \multicolumn{1}{c}{$t_k$} 
& \multicolumn{1}{c}{$E_k$} & \multicolumn{1}{c}{$t_k$} 
& \multicolumn{1}{c}{$E_k$} & \multicolumn{1}{c}{$t_k$} 
& \multicolumn{1}{c}{$E_k$} & \multicolumn{1}{c}{$t_k$} 
& \multicolumn{1}{c}{$E_k$} & \multicolumn{1}{c}{$t_k$} 
& \multicolumn{1}{c}{$E_k$} & \multicolumn{1}{c}{$t_k$} \\
\midrule
2	&	3.68403	&	0.00	&	0.02	&	0.00	&	1.08	&	0.00	&	0.75	&	0.05	&	0.19	&	0.01	&	3.35	&	0.30	&	0.06	\\
3	&	2.53240	&	0.00	&	0.11	&	0.00	&	1.48	&	0.00	&	1.41	&	0.12	&	0.35	&	0.02	&	3.26	&	2.17	&	0.05	\\
4	&	1.73600	&	0.00	&	0.17	&	0.00	&	1.81	&	0.00	&	1.81	&	0.16	&	0.50	&	0.02	&	3.40	&	8.27	&	0.05	\\
5	&	1.32707*	&	0.00	&	0.23	&	0.00	&	2.11	&	0.00	&	2.23	&	0.17	&	0.64	&	0.03	&	3.26	&	3.46	&	0.05	\\
10	&	0.64490*	&	0.00	&	0.55	&	1.41	&	2.98	&	0.62	&	5.02	&	0.75	&	1.39	&	1.18	&	3.31	&	8.61	&	0.05	\\
15	&	0.43136	&	0.00	&	0.91	&	0.24	&	3.61	&	0.25	&	9.95	&	1.53	&	2.07	&	1.12	&	3.27	&	8.30	&	0.05	\\
20	&	0.32177	&	0.92	&	1.44	&	0.24	&	4.30	&	0.24	&	15.39	&	2.14	&	3.11	&	1.17	&	3.26	&	6.00	&	0.05	\\
25	&	0.25308	&	1.04	&	1.95	&	0.48	&	4.94	&	0.47	&	24.52	&	2.86	&	4.16	&	0.71	&	3.26	&	7.75	&	0.06	\\

\midrule
$E_\text{aver}$	&		&	0.25	&		&	0.30	&		&	0.20	&		&	0.97	&		&	0.53	&		&	5.61	&		\\
$t_\text{init}$	&		&		&	0.05	&		&	0.05	&		&	0.03	&		&	0.05	&		&	0.01	&		&	0.00	\\
$t_\text{total}$	&		&		&	2.00	&		&	4.98	&		&	24.55	&		&	4.20	&		&	26.38	&		&	0.42	\\

\bottomrule
\end{tabular}
\label{taulukko_d15112}
} 
\end{table}

\begin{table}[ht!]
\centering
\caption{Summary of the results with Online News Popularity ($\times 10^{14}$). Dimensions: $m$ = 39,644, $n$ = 58.} \vspace{0.25cm}
\resizebox{\textwidth}{!}{ 
\begin{tabular}{
    c c 
    c c
    c c
    c c
    c c
    c c
    c c
}
\toprule
$k$ & $f_\text{best}$ 
& \multicolumn{2}{c}{\textsc{Clust-Splitter}} 
& \multicolumn{2}{c}{\textsc{LMBM-Clust}} 
& \multicolumn{2}{c}{\textsc{DC-Clust}}
& \multicolumn{2}{c}{\textsc{Big-Clust}}
& \multicolumn{2}{c}{\textsc{Big-Means}}
& \multicolumn{2}{c}{\textsc{MiniBatchKMeans}} \\
\cmidrule(lr){3-4} \cmidrule(lr){5-6} \cmidrule(lr){7-8} \cmidrule(lr){9-10} \cmidrule(lr){11-12} \cmidrule(lr){13-14}
& & \multicolumn{1}{c}{$E_k$} & \multicolumn{1}{c}{$t_k$} 
& \multicolumn{1}{c}{$E_k$} & \multicolumn{1}{c}{$t_k$} 
& \multicolumn{1}{c}{$E_k$} & \multicolumn{1}{c}{$t_k$} 
& \multicolumn{1}{c}{$E_k$} & \multicolumn{1}{c}{$t_k$} 
& \multicolumn{1}{c}{$E_k$} & \multicolumn{1}{c}{$t_k$} 
& \multicolumn{1}{c}{$E_k$} & \multicolumn{1}{c}{$t_k$} \\
\midrule
2	&	9.53913*	&	0.00	&	11.28	&	0.00	&	1.86	&	0.00	&	18.08	&	0.02	&	0.62	&	0.02	&	4.26	&	2.13	&	0.10	\\
3	&	5.91077	&	0.00	&	16.89	&	0.00	&	26.56	&	0.00	&	45.73	&	0.04	&	1.07	&	0.09	&	4.25	&	3.32	&	0.05	\\
4	&	4.30793	&	7.66	&	22.13	&	7.66	&	27.28	&	0.00	&	76.89	&	7.70	&	1.55	&	3.92	&	4.26	&	2.16	&	0.05	\\
5	&	3.09885	&	0.00	&	27.19	&	0.00	&	39.12	&	0.00	&	103.64	&	0.07	&	2.07	&	1.88	&	4.26	&	14.62	&	0.05	\\
10	&	1.17247	&	2.58	&	52.80	&	2.57	&	54.06	&	0.00	&	303.48	&	8.29	&	4.39	&	5.85	&	4.27	&	35.42	&	0.06	\\
15	&	0.77637	&	0.00	&	97.92	&	14.77	&	65.48	&	0.00	&	577.20	&	21.36	&	7.40	&	6.55	&	4.50	&	38.53	&	0.07	\\
20	&	0.59809	&	1.93	&	131.52	&	19.49	&	80.97	&	0.00	&	990.88	&	14.24	&	11.18	&	4.37	&	4.26	&	25.08	&	0.07	\\
25	&	0.49616	&	1.55	&	164.88	&	7.03	&	93.77	&	0.00	&	1443.78	&	9.53	&	15.46	&	5.23	&	4.28	&	32.91	&	0.08	\\

\midrule
$E_\text{aver}$	&		&	1.72	&		&	6.44	&		&	0.00	&		&	7.66	&		&	3.49	&		&	19.27	&		\\
$t_\text{init}$	&		&		&	2.95	&		&	2.66	&		&	2.69	&		&	2.70	&		&	0.49	&		&	0.50	\\
$t_\text{total}$	&		&		&	167.83	&		&	96.42	&		&	1446.47	&		&	18.16	&		&	34.83	&		&	1.04	\\

\bottomrule
\end{tabular}
\label{taulukko_onlinenews}
} 
\end{table}

\begin{table}[ht!]
\centering
\caption{Summary of the results with KEGG Metabolic ($\times 10^8$). Dimensions: $m$ = 53,413, $n$ = 20.} \vspace{0.25cm}
\resizebox{\textwidth}{!}{ 
\begin{tabular}{
    c c 
    c c
    c c
    c c
    c c
    c c
    c c
}
\toprule
$k$ & $f_\text{best}$ 
& \multicolumn{2}{c}{\textsc{Clust-Splitter}} 
& \multicolumn{2}{c}{\textsc{LMBM-Clust}} 
& \multicolumn{2}{c}{\textsc{DC-Clust}}
& \multicolumn{2}{c}{\textsc{Big-Clust}}
& \multicolumn{2}{c}{\textsc{Big-Means}}
& \multicolumn{2}{c}{\textsc{MiniBatchKMeans}} \\
\cmidrule(lr){3-4} \cmidrule(lr){5-6} \cmidrule(lr){7-8} \cmidrule(lr){9-10} \cmidrule(lr){11-12} \cmidrule(lr){13-14}
& & \multicolumn{1}{c}{$E_k$} & \multicolumn{1}{c}{$t_k$} 
& \multicolumn{1}{c}{$E_k$} & \multicolumn{1}{c}{$t_k$} 
& \multicolumn{1}{c}{$E_k$} & \multicolumn{1}{c}{$t_k$} 
& \multicolumn{1}{c}{$E_k$} & \multicolumn{1}{c}{$t_k$} 
& \multicolumn{1}{c}{$E_k$} & \multicolumn{1}{c}{$t_k$} 
& \multicolumn{1}{c}{$E_k$} & \multicolumn{1}{c}{$t_k$} \\
\midrule
2	&	11.38530	&	0.00	&	0.22	&	0.00	&	0.25	&	0.00	&	2.81	&	0.01	&	0.30	&	5.89	&	4.27	&	23.66	&	0.06	\\
3	&	4.90060*	&	0.00	&	0.61	&	0.00	&	0.53	&	0.00	&	8.05	&	0.21	&	0.49	&	0.69	&	4.27	&	160.13	&	0.03	\\
4	&	2.72950	&	0.01	&	1.00	&	0.00	&	0.91	&	0.00	&	17.42	&	2.04	&	0.77	&	1.10	&	4.27	&	316.62	&	0.04	\\
5	&	1.88367	&	0.00	&	2.02	&	0.00	&	1.47	&	0.00	&	31.30	&	2.24	&	1.05	&	1.70	&	4.27	&	482.70	&	0.03	\\
10	&	0.60513	&	5.17	&	7.05	&	4.96	&	4.05	&	5.00	&	139.17	&	9.65	&	2.34	&	38.11	&	4.27	&	1516.83	&	0.03	\\
15	&	0.34940*	&	0.00	&	20.02	&	2.01	&	8.45	&	0.93	&	308.75	&	9.24	&	3.57	&	86.29	&	4.39	&	2542.99	&	0.05	\\
20	&	0.25027	&	0.47	&	35.33	&	7.72	&	12.33	&	0.31	&	477.13	&	10.55	&	4.92	&	88.20	&	4.28	&	3679.29	&	0.04	\\
25	&	0.19253*	&	0.00	&	67.94	&	2.83	&	17.69	&	2.11	&	692.36	&	9.23	&	6.36	&	120.35	&	4.28	&	4692.85	&	0.06	\\

\midrule
$E_\text{aver}$	&		&	0.71	&		&	2.19	&		&	1.04	&		&	5.40	&		&	42.79	&		&	1676.89	&		\\
$t_\text{init}$	&		&		&	1.41	&		&	1.28	&		&	1.30	&		&	1.36	&		&	0.24	&		&	0.27	\\
$t_\text{total}$	&		&		&	69.34	&		&	18.97	&		&	693.66	&		&	7.72	&		&	34.54	&		&	0.61	\\

\bottomrule
\end{tabular}
\label{taulukko_kegg}
} 
\end{table}

\begin{table}[ht!]
\centering
\caption{Summary of the results with Shuttle Control ($\times 10^8$). Dimensions: $m$ = 58,000, $n$ = 9.} \vspace{0.25cm}
\resizebox{\textwidth}{!}{ 
\begin{tabular}{
    c c 
    c c
    c c
    c c
    c c
    c c
    c c
}
\toprule
$k$ & $f_\text{best}$ 
& \multicolumn{2}{c}{\textsc{Clust-Splitter}} 
& \multicolumn{2}{c}{\textsc{LMBM-Clust}} 
& \multicolumn{2}{c}{\textsc{DC-Clust}}
& \multicolumn{2}{c}{\textsc{Big-Clust}}
& \multicolumn{2}{c}{\textsc{Big-Means}}
& \multicolumn{2}{c}{\textsc{MiniBatchKMeans}} \\
\cmidrule(lr){3-4} \cmidrule(lr){5-6} \cmidrule(lr){7-8} \cmidrule(lr){9-10} \cmidrule(lr){11-12} \cmidrule(lr){13-14}
& & \multicolumn{1}{c}{$E_k$} & \multicolumn{1}{c}{$t_k$} 
& \multicolumn{1}{c}{$E_k$} & \multicolumn{1}{c}{$t_k$} 
& \multicolumn{1}{c}{$E_k$} & \multicolumn{1}{c}{$t_k$} 
& \multicolumn{1}{c}{$E_k$} & \multicolumn{1}{c}{$t_k$} 
& \multicolumn{1}{c}{$E_k$} & \multicolumn{1}{c}{$t_k$} 
& \multicolumn{1}{c}{$E_k$} & \multicolumn{1}{c}{$t_k$} \\
\midrule
2	&	21.34329	&	0.00	&	0.13	&	0.00	&	0.23	&	0.00	&	0.47	&	0.00	&	0.36	&	9.80	&	3.26	&	52.07	&	0.03	\\
3	&	10.85415	&	0.00	&	0.36	&	0.00	&	0.39	&	0.00	&	1.47	&	0.01	&	0.51	&	22.85	&	3.31	&	196.60	&	0.04	\\
4	&	8.86910	&	5.28	&	0.52	&	0.00	&	0.55	&	0.39	&	3.66	&	3.33	&	0.74	&	17.39	&	3.25	&	260.32	&	0.03	\\
5	&	7.24479	&	11.00	&	1.73	&	0.09	&	0.70	&	0.44	&	7.81	&	2.86	&	0.96	&	39.90	&	3.26	&	333.09	&	0.05	\\
10	&	2.83216	&	61.76	&	4.55	&	0.55	&	1.95	&	0.24	&	45.66	&	4.31	&	2.34	&	64.84	&	3.26	&	990.79	&	0.03	\\
15	&	1.53154	&	128.43	&	10.27	&	0.02	&	3.28	&	3.59	&	192.28	&	12.18	&	3.64	&	114.08	&	3.26	&	1860.54	&	0.05	\\
20	&	1.05032	&	196.01	&	16.81	&	0.96	&	5.33	&	0.00	&	292.42	&	10.20	&	5.00	&	190.94	&	3.27	&	2677.05	&	0.04	\\
25	&	0.77978	&	275.97	&	23.77	&	0.00	&	8.61	&	3.55	&	414.50	&	10.57	&	6.59	&	140.21	&	3.27	&	3556.60	&	0.03	\\

\midrule
$E_\text{aver}$	&		&	84.80	&		&	0.20	&		&	1.03	&		&	5.43	&		&	75.00	&		&	1240.88	&		\\
$t_\text{init}$	&		&		&	0.63	&		&	0.58	&		&	0.58	&		&	0.60	&		&	0.18	&		&	0.17	\\
$t_\text{total}$	&		&		&	24.39	&		&	9.19	&		&	415.08	&		&	7.19	&		&	26.32	&		&	0.47	\\

\bottomrule
\end{tabular}
\label{taulukko_shuttle}
} 
\end{table}

\begin{table}[ht!]
\centering
\caption{Summary of the results with Sensorless Drive Diagnosis ($\times 10^7$). Dimensions: $m$ = 58,509, $n$ = 48.} \vspace{0.25cm}
\resizebox{\textwidth}{!}{ 
\begin{tabular}{
    c c 
    c c
    c c
    c c
    c c
    c c
    c c
}
\toprule
$k$ & $f_\text{best}$ 
& \multicolumn{2}{c}{\textsc{Clust-Splitter}} 
& \multicolumn{2}{c}{\textsc{LMBM-Clust}} 
& \multicolumn{2}{c}{\textsc{DC-Clust}}
& \multicolumn{2}{c}{\textsc{Big-Clust}}
& \multicolumn{2}{c}{\textsc{Big-Means}}
& \multicolumn{2}{c}{\textsc{MiniBatchKMeans}} \\
\cmidrule(lr){3-4} \cmidrule(lr){5-6} \cmidrule(lr){7-8} \cmidrule(lr){9-10} \cmidrule(lr){11-12} \cmidrule(lr){13-14}
& & \multicolumn{1}{c}{$E_k$} & \multicolumn{1}{c}{$t_k$} 
& \multicolumn{1}{c}{$E_k$} & \multicolumn{1}{c}{$t_k$} 
& \multicolumn{1}{c}{$E_k$} & \multicolumn{1}{c}{$t_k$} 
& \multicolumn{1}{c}{$E_k$} & \multicolumn{1}{c}{$t_k$} 
& \multicolumn{1}{c}{$E_k$} & \multicolumn{1}{c}{$t_k$} 
& \multicolumn{1}{c}{$E_k$} & \multicolumn{1}{c}{$t_k$} \\
\midrule
2	&	3.88116	&	1.51	&	0.42	&	1.51	&	0.44	&	1.51	&	7.50	&	1.81	&	0.66	&	57.81	&	4.59	&	101.67	&	0.04	\\
3	&	2.91313	&	26.29	&	0.92	&	2.01	&	0.62	&	3.93	&	31.72	&	7.26	&	0.95	&	57.04	&	4.32	&	158.89	&	0.03	\\
4	&	2.26160	&	33.94	&	3.66	&	2.59	&	1.33	&	4.94	&	62.64	&	11.97	&	1.32	&	83.97	&	4.27	&	221.94	&	0.03	\\
5	&	1.93651	&	39.65	&	5.53	&	10.52	&	1.75	&	5.55	&	110.50	&	9.47	&	1.77	&	92.16	&	4.33	&	271.89	&	0.04	\\
10	&	0.96090	&	118.11	&	29.19	&	10.50	&	5.50	&	7.87	&	428.95	&	8.20	&	3.51	&	111.71	&	4.28	&	603.18	&	0.04	\\
15	&	0.62816	&	206.18	&	58.34	&	9.31	&	20.25	&	10.91	&	878.84	&	5.12	&	6.76	&	141.36	&	4.28	&	944.30	&	0.04	\\
20	&	0.49884	&	270.39	&	90.19	&	14.28	&	30.81	&	11.79	&	1346.34	&	6.93	&	9.96	&	269.20	&	4.32	&	1201.56	&	0.04	\\
25	&	0.42225	&	324.60	&	132.53	&	15.89	&	44.22	&	12.82	&	1922.28	&	8.20	&	13.56	&	286.79	&	4.43	&	1420.27	&	0.04	\\

\midrule
$E_\text{aver}$	&		&	127.58	&		&	8.33	&		&	7.41	&		&	7.37	&		&	137.50	&		&	615.46	&		\\
$t_\text{init}$	&		&		&	4.27	&		&	3.89	&		&	3.95	&		&	3.91	&		&	0.77	&		&	0.75	\\
$t_\text{total}$	&		&		&	136.80	&		&	48.11	&		&	1926.23	&		&	17.47	&		&	35.58	&		&	1.05	\\

\bottomrule
\end{tabular}
\label{taulukko_sensorless}
} 
\end{table}

\begin{figure}[ht!]
    \centering
    \includegraphics[width=\textwidth]{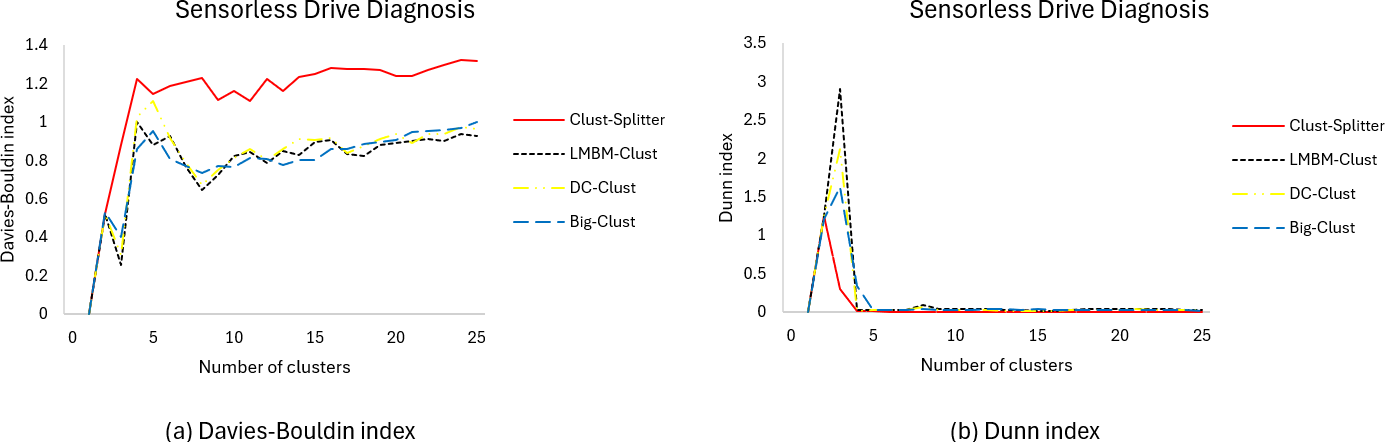}
    \caption{Sensorless Drive Diagnosis: Davies-Bouldin and Dunn validity indices vs. number of clusters.}
    \label{index_snesorless}
\end{figure}

\begin{table}[ht!]
\centering
\caption{Summary of the results with MFCCs for Speech Emotion Recognition ($\times 10^8$). Dimensions: $m$ = 85,134, $n$ = 58.} \vspace{0.25cm}
\resizebox{\textwidth}{!}{ 
\begin{tabular}{
    c c 
    c c
    c c
    c c
    c c
    c c
    c c
}
\toprule
$k$ & $f_\text{best}$ 
& \multicolumn{2}{c}{\textsc{Clust-Splitter}} 
& \multicolumn{2}{c}{\textsc{LMBM-Clust}} 
& \multicolumn{2}{c}{\textsc{DC-Clust}}
& \multicolumn{2}{c}{\textsc{Big-Clust}}
& \multicolumn{2}{c}{\textsc{Big-Means}}
& \multicolumn{2}{c}{\textsc{MiniBatchKMeans}} \\
\cmidrule(lr){3-4} \cmidrule(lr){5-6} \cmidrule(lr){7-8} \cmidrule(lr){9-10} \cmidrule(lr){11-12} \cmidrule(lr){13-14}
& & \multicolumn{1}{c}{$E_k$} & \multicolumn{1}{c}{$t_k$} 
& \multicolumn{1}{c}{$E_k$} & \multicolumn{1}{c}{$t_k$} 
& \multicolumn{1}{c}{$E_k$} & \multicolumn{1}{c}{$t_k$} 
& \multicolumn{1}{c}{$E_k$} & \multicolumn{1}{c}{$t_k$} 
& \multicolumn{1}{c}{$E_k$} & \multicolumn{1}{c}{$t_k$} 
& \multicolumn{1}{c}{$E_k$} & \multicolumn{1}{c}{$t_k$} \\
\midrule
2	&	7.45130	&	0.00	&	1.81	&	0.00	&	6.45	&	0.00	&	125.27	&	0.84	&	1.60	&	0.01	&	4.29	&	3.25	&	0.12	\\
3	&	5.02150	&	0.00	&	6.03	&	0.00	&	8.41	&	0.00	&	250.03	&	0.09	&	2.54	&	0.01	&	4.29	&	5.53	&	0.14	\\
4	&	4.16900	&	1.67	&	10.44	&	0.00	&	11.03	&	0.00	&	413.13	&	0.15	&	4.21	&	0.51	&	4.41	&	1.59	&	0.10	\\
5	&	3.45592	&	0.00	&	13.16	&	0.00	&	15.66	&	0.00	&	574.66	&	0.25	&	6.11	&	0.01	&	4.35	&	5.01	&	0.13	\\
10	&	2.17618	&	2.47	&	33.11	&	0.00	&	29.59	&	0.00	&	1464.70	&	1.13	&	12.28	&	0.98	&	4.29	&	2.18	&	0.14	\\
15	&	1.76044	&	0.00	&	71.27	&	1.19	&	47.28	&	0.00	&	2478.06	&	1.93	&	19.57	&	0.87	&	4.31	&	3.41	&	0.12	\\
20	&	1.53799*	&	0.00	&	121.84	&	2.57	&	63.44	&	0.31	&	3627.80	&	3.65	&	26.23	&	1.01	&	4.33	&	3.69	&	0.17	\\
25	&	1.41090	&	0.93	&	174.50	&	0.74	&	84.98	&	0.61	&	4983.70	&	2.61	&	34.82	&	1.01	&	4.31	&	3.59	&	0.14	\\

\midrule
$E_\text{aver}$	&		&	0.63	&		&	0.56	&		&	0.12	&		&	1.33	&		&	0.55	&		&	3.53	&		\\
$t_\text{init}$	&		&		&	7.23	&		&	6.94	&		&	6.94	&		&	6.95	&		&	2.08	&		&	2.05	\\
$t_\text{total}$	&		&		&	181.73	&		&	91.92	&		&	4990.64	&		&	41.77	&		&	36.65	&		&	3.11	\\

\bottomrule
\end{tabular}
\label{taulukko_mfcc}
} 
\end{table}

\begin{table}[ht!]
\centering
\caption{Summary of the results with Pla85900 ($\times 10^{15}$). Dimensions: $m$ = 85,900, $n$ = 2.} \vspace{0.25cm}
\resizebox{\textwidth}{!}{ 
\begin{tabular}{
    c c 
    c c
    c c
    c c
    c c
    c c
    c c
}
\toprule
$k$ & $f_\text{best}$ 
& \multicolumn{2}{c}{\textsc{Clust-Splitter}} 
& \multicolumn{2}{c}{\textsc{LMBM-Clust}} 
& \multicolumn{2}{c}{\textsc{DC-Clust}}
& \multicolumn{2}{c}{\textsc{Big-Clust}}
& \multicolumn{2}{c}{\textsc{Big-Means}}
& \multicolumn{2}{c}{\textsc{MiniBatchKMeans}} \\
\cmidrule(lr){3-4} \cmidrule(lr){5-6} \cmidrule(lr){7-8} \cmidrule(lr){9-10} \cmidrule(lr){11-12} \cmidrule(lr){13-14}
& & \multicolumn{1}{c}{$E_k$} & \multicolumn{1}{c}{$t_k$} 
& \multicolumn{1}{c}{$E_k$} & \multicolumn{1}{c}{$t_k$} 
& \multicolumn{1}{c}{$E_k$} & \multicolumn{1}{c}{$t_k$} 
& \multicolumn{1}{c}{$E_k$} & \multicolumn{1}{c}{$t_k$} 
& \multicolumn{1}{c}{$E_k$} & \multicolumn{1}{c}{$t_k$} 
& \multicolumn{1}{c}{$E_k$} & \multicolumn{1}{c}{$t_k$} \\
\midrule
2	&	3.74908	&	1.44	&	1.53	&	1.44	&	3.30	&	0.00	&	7.72	&	1.80	&	0.50	&	1.01	&	3.27	&	1.48	&	0.12	\\
3	&	2.28057	&	0.00	&	2.67	&	0.00	&	4.69	&	0.00	&	15.09	&	0.02	&	0.84	&	0.01	&	3.26	&	2.43	&	0.12	\\
4	&	1.59308*	&	0.00	&	3.19	&	0.00	&	5.88	&	0.00	&	22.13	&	0.03	&	1.17	&	0.01	&	3.33	&	3.99	&	0.16	\\
5	&	1.33972	&	0.81	&	3.77	&	2.77	&	6.91	&	0.00	&	31.02	&	0.03	&	1.48	&	0.81	&	3.27	&	1.61	&	0.13	\\
10	&	0.68294	&	0.55	&	7.06	&	0.40	&	16.36	&	0.00	&	84.81	&	1.49	&	3.03	&	0.42	&	3.27	&	3.98	&	0.12	\\
15	&	0.46029	&	0.16	&	11.69	&	0.98	&	22.52	&	0.48	&	143.27	&	1.26	&	4.78	&	0.36	&	3.27	&	4.95	&	0.14	\\
20	&	0.34988	&	1.60	&	23.23	&	0.94	&	28.34	&	0.52	&	203.89	&	0.76	&	6.82	&	0.52	&	3.27	&	5.65	&	0.15	\\
25	&	0.28259	&	0.29	&	34.00	&	0.18	&	37.34	&	0.02	&	274.70	&	1.23	&	9.17	&	0.70	&	3.27	&	4.60	&	0.11	\\

\midrule
$E_\text{aver}$	&		&	0.61	&		&	0.84	&		&	0.13	&		&	0.83	&		&	0.48	&		&	3.59	&		\\
$t_\text{init}$	&		&		&	0.27	&		&	0.25	&		&	0.25	&		&	0.27	&		&	0.08	&		&	0.08	\\
$t_\text{total}$	&		&		&	34.27	&		&	37.59	&		&	274.95	&		&	9.43	&		&	26.29	&		&	1.13	\\

\bottomrule
\end{tabular}
\label{taulukko_pla85900}
} 
\end{table}

\begin{figure}[ht!]
    \centering
    \includegraphics[width=\textwidth]{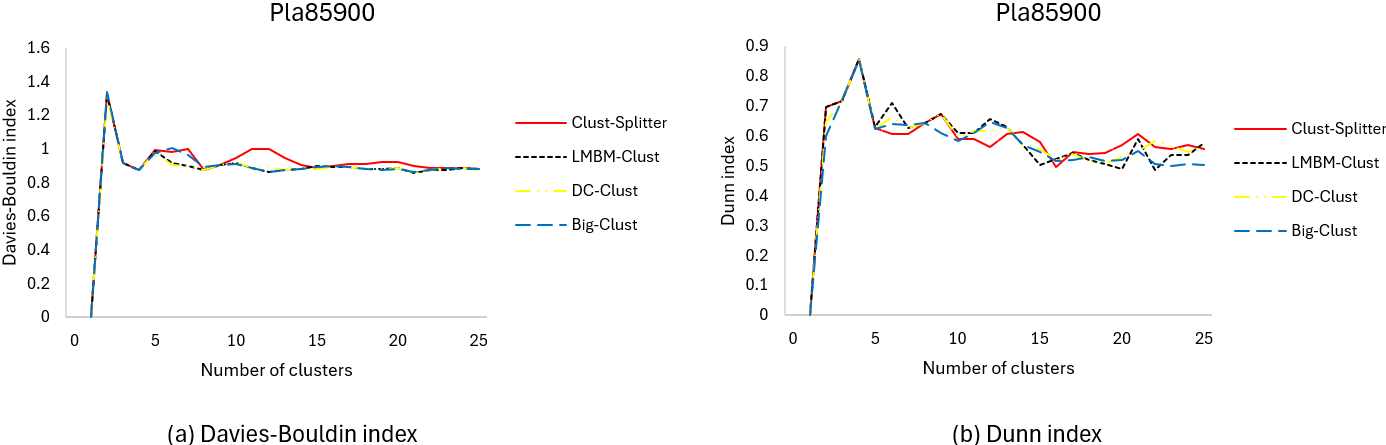}
    \caption{Pla85900: Davies-Bouldin and Dunn validity indices vs. number of clusters.}
    \label{index_pla85900}
\end{figure}

\begin{table}[ht!]
\centering
\caption{Summary of the results with Music Analysis ($\times 10^{11}$). Dimensions: $m$ = 106,574, $n$ = 518.} \vspace{0.25cm}
\resizebox{\textwidth}{!}{ 
\begin{tabular}{
    c c 
    c c
    c c
    c c
    c c
    c c
    c c
}
\toprule
$k$ & $f_\text{best}$ 
& \multicolumn{2}{c}{\textsc{Clust-Splitter}} 
& \multicolumn{2}{c}{\textsc{LMBM-Clust}} 
& \multicolumn{2}{c}{\textsc{DC-Clust}}
& \multicolumn{2}{c}{\textsc{Big-Clust}}
& \multicolumn{2}{c}{\textsc{Big-Means}}
& \multicolumn{2}{c}{\textsc{MiniBatchKMeans}} \\
\cmidrule(lr){3-4} \cmidrule(lr){5-6} \cmidrule(lr){7-8} \cmidrule(lr){9-10} \cmidrule(lr){11-12} \cmidrule(lr){13-14}
& & \multicolumn{1}{c}{$E_k$} & \multicolumn{1}{c}{$t_k$} 
& \multicolumn{1}{c}{$E_k$} & \multicolumn{1}{c}{$t_k$} 
& \multicolumn{1}{c}{$E_k$} & \multicolumn{1}{c}{$t_k$} 
& \multicolumn{1}{c}{$E_k$} & \multicolumn{1}{c}{$t_k$} 
& \multicolumn{1}{c}{$E_k$} & \multicolumn{1}{c}{$t_k$} 
& \multicolumn{1}{c}{$E_k$} & \multicolumn{1}{c}{$t_k$} \\
\midrule
2	&	5.00470	&	0.00	&	11.77	&	0.00	&	16.48	&	0.00	&	4137.12	&	0.03	&	7.96	&	0.00	&	3.36	&	0.21	&	0.36	\\
3	&	3.83746	&	0.00	&	50.09	&	0.00	&	31.08	&	0.00	&	7559.89	&	0.09	&	11.72	&	4.54	&	3.35	&	5.36	&	0.30	\\
4	&	3.11830	&	0.00	&	111.72	&	0.00	&	63.42	&	0.00	&	12,042.67	&	0.13	&	17.83	&	3.95	&	3.38	&	7.98	&	0.53	\\
5	&	2.74247	&	0.00	&	138.00	&	0.00	&	85.59	&	0.00	&	16,183.20	&	0.20	&	22.71	&	0.91	&	3.36	&	5.47	&	0.50	\\
10	&	1.87257	&	1.70	&	448.25	&	0.00	&	195.33	&	0.00	&	41,252.87	&	0.79	&	56.11	&	1.00	&	3.60	&	4.75	&	0.34	\\
15	&	1.54420	&	0.42	&	1049.08	&	0.52	&	392.30	&	0.42	&	68,087.94	&	1.48	&	107.47	&	0.88	&	4.37	&	3.43	&	0.41	\\
20	&	1.35320	&	0.10	&	1824.34	&	0.08	&	628.92	&	$-$	&	$-$	&	2.45	&	161.17	&	0.59	&	5.98	&	3.88	&	0.41	\\
25	&	1.22620	&	1.33	&	2701.72	&	1.64	&	908.87	&	$-$	&	$-$	&	3.12	&	224.96	&	0.73	&	6.44	&	3.94	&	0.44	\\

\midrule
$E_\text{aver}$	&		&	0.44	&		&	0.28	&		&	0.07	&		&	1.03	&		&	1.57	&		&	4.38	&		\\
$t_\text{init}$	&		&		&	78.52	&		&	74.39	&		&	74.41	&		&	74.19	&		&	14.14	&		&	14.14	\\
$t_\text{total}$	&		&		&	2780.23	&		&	983.27	&		&	68,162.34	&		&	299.15	&		&	47.99	&		&	17.42	\\

\bottomrule
\end{tabular}
\label{taulukko_music}
} 
\end{table}

\begin{figure}[ht!]
    \centering
    \includegraphics[width=\textwidth]{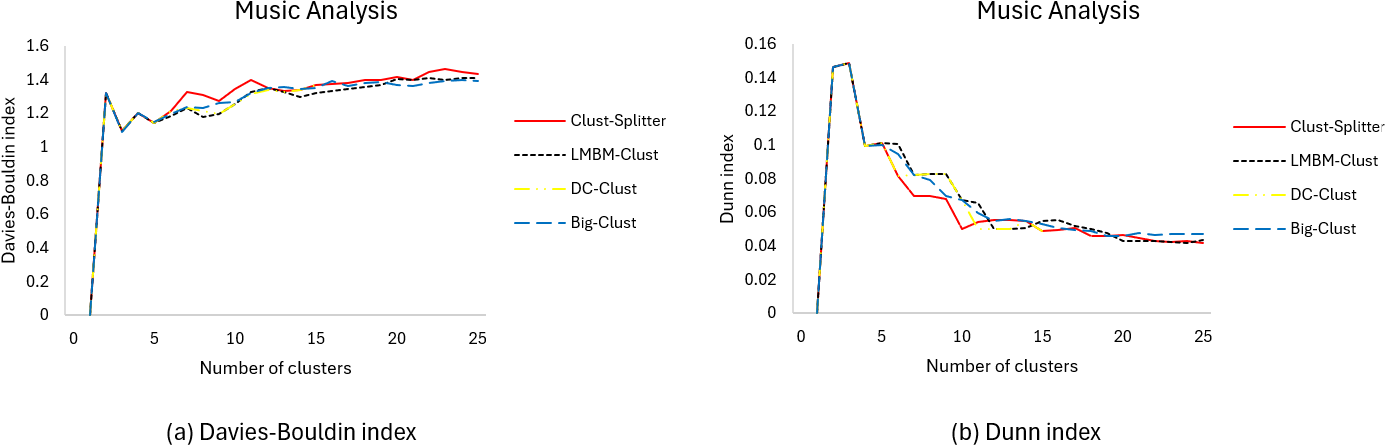}
    \caption{Music Analysis: Davies-Bouldin and Dunn validity indices vs. number of clusters.}
    \label{index_music}
\end{figure}

\begin{table}[ht!]
\centering
\caption{Summary of the results with MiniBooNE Particle Identification ($\times 10^{10}$). Dimensions: $m$ = 130,064, $n$ = 50.} \vspace{0.25cm}
\resizebox{\textwidth}{!}{ 
\begin{tabular}{
    c c 
    c c
    c c
    c c
    c c
    c c
    c c
}
\toprule
$k$ & $f_\text{best}$ 
& \multicolumn{2}{c}{\textsc{Clust-Splitter}} 
& \multicolumn{2}{c}{\textsc{LMBM-Clust}} 
& \multicolumn{2}{c}{\textsc{DC-Clust}}
& \multicolumn{2}{c}{\textsc{Big-Clust}}
& \multicolumn{2}{c}{\textsc{Big-Means}}
& \multicolumn{2}{c}{\textsc{MiniBatchKMeans}} \\
\cmidrule(lr){3-4} \cmidrule(lr){5-6} \cmidrule(lr){7-8} \cmidrule(lr){9-10} \cmidrule(lr){11-12} \cmidrule(lr){13-14}
& & \multicolumn{1}{c}{$E_k$} & \multicolumn{1}{c}{$t_k$} 
& \multicolumn{1}{c}{$E_k$} & \multicolumn{1}{c}{$t_k$} 
& \multicolumn{1}{c}{$E_k$} & \multicolumn{1}{c}{$t_k$} 
& \multicolumn{1}{c}{$E_k$} & \multicolumn{1}{c}{$t_k$} 
& \multicolumn{1}{c}{$E_k$} & \multicolumn{1}{c}{$t_k$} 
& \multicolumn{1}{c}{$E_k$} & \multicolumn{1}{c}{$t_k$} \\
\midrule
2	&	8.92236	&	252,620.64	&	26.42	&	0.00	&	0.98	&	0.00	&	4.41	&	0.00	&	1.15	&	286,911.70	&	4.27	&	286,901.67	&	0.07	\\
3	&	5.22601	&	404,522.85	&	55.98	&	21.68	&	1.70	&	0.00	&	102.80	&	21.69	&	1.97	&	391,916.32	&	4.28	&	489,605.83	&	0.07	\\
4	&	2.70080	&	730,252.51	&	94.44	&	0.00	&	3.05	&	0.00	&	189.31	&	0.04	&	2.99	&	947,964.98	&	4.46	&	946,975.49	&	0.07	\\
5	&	1.82252	&	1,006,752.24	&	131.94	&	0.00	&	6.84	&	0.00	&	441.72	&	0.12	&	5.71	&	1,264,307.71	&	4.42	&	1,404,755.54	&	0.08	\\
10	&	0.90920	&	1,348,100.88	&	487.41	&	1.63	&	31.06	&	1.65	&	2025.89	&	2.37	&	15.03	&	2,815,972.07	&	4.31	&	2,813,472.85	&	0.07	\\
15	&	0.63506	&	1,165,473.95	&	1059.80	&	0.00	&	59.73	&	0.02	&	3897.16	&	1.47	&	22.30	&	4,031,566.07	&	4.33	&	4,029,153.17	&	0.09	\\
20	&	0.50863	&	739,691.13	&	1731.14	&	1.17	&	99.87	&	0.37	&	6035.58	&	2.30	&	31.07	&	4,530,328.25	&	4.41	&	5,006,683.99	&	0.09	\\
25	&	0.44425	&	309,221.62	&	2535.77	&	1.12	&	141.33	&	0.03	&	8274.08	&	1.76	&	41.59	&	5,763,170.88	&	4.41	&	5,754,541.94	&	0.08	\\

\midrule
$E_\text{aver}$	&		&	744,579.48	&		&	3.20	&		&	0.26	&		&	3.72	&		&	2,504,017.25	&		&	2,591,511.31	&		\\
$t_\text{init}$	&		&		&	9.42	&		&	9.08	&		&	9.08	&		&	9.57	&		&	1.83	&		&	1.86	\\
$t_\text{total}$	&		&		&	2545.19	&		&	150.41	&		&	8283.16	&		&	131.36	&		&	36.72	&		&	2.47	\\

\bottomrule
\end{tabular}
\label{taulukko_miniboone}
} 
\end{table}

\begin{figure}[ht!]
    \centering
    \includegraphics[width=\textwidth]{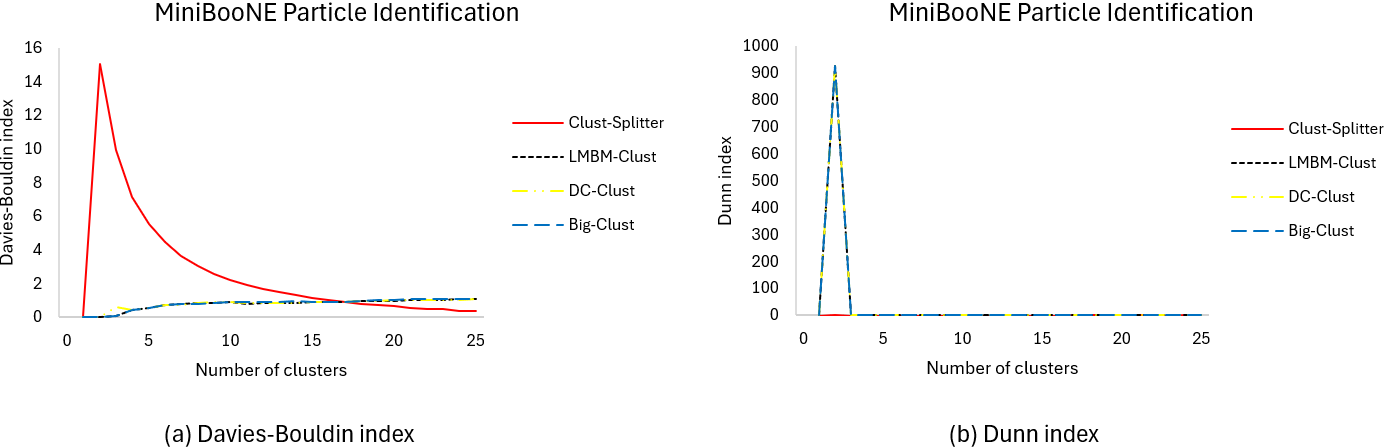}
    \caption{MiniBooNE Particle Identification: Davies-Bouldin and Dunn validity indices vs. number of clusters.}
    \label{index_miniboone}
\end{figure}

\begin{table}[ht!]
\centering
\caption{Summary of the results with Protein Homology ($\times 10^{11}$). Dimensions: $m$ = 145,751, $n$ = 74.} \vspace{0.25cm}
\resizebox{\textwidth}{!}{ 
\begin{tabular}{
    c c 
    c c
    c c
    c c
    c c
    c c
    c c
}
\toprule
$k$ & $f_\text{best}$ 
& \multicolumn{2}{c}{\textsc{Clust-Splitter}} 
& \multicolumn{2}{c}{\textsc{LMBM-Clust}} 
& \multicolumn{2}{c}{\textsc{DC-Clust}}
& \multicolumn{2}{c}{\textsc{Big-Clust}}
& \multicolumn{2}{c}{\textsc{Big-Means}}
& \multicolumn{2}{c}{\textsc{MiniBatchKMeans}} \\
\cmidrule(lr){3-4} \cmidrule(lr){5-6} \cmidrule(lr){7-8} \cmidrule(lr){9-10} \cmidrule(lr){11-12} \cmidrule(lr){13-14}
& & \multicolumn{1}{c}{$E_k$} & \multicolumn{1}{c}{$t_k$} 
& \multicolumn{1}{c}{$E_k$} & \multicolumn{1}{c}{$t_k$} 
& \multicolumn{1}{c}{$E_k$} & \multicolumn{1}{c}{$t_k$} 
& \multicolumn{1}{c}{$E_k$} & \multicolumn{1}{c}{$t_k$} 
& \multicolumn{1}{c}{$E_k$} & \multicolumn{1}{c}{$t_k$} 
& \multicolumn{1}{c}{$E_k$} & \multicolumn{1}{c}{$t_k$} \\
\midrule
2	&	15.20430	&	1.82	&	2.55	&	0.00	&	1.52	&	0.00	&	167.09	&	0.03	&	2.36	&	1.21	&	26.30	&	2.67	&	0.13	\\
3	&	8.07129	&	0.00	&	7.56	&	0.00	&	5.13	&	0.00	&	457.25	&	0.15	&	3.14	&	1.04	&	26.29	&	63.92	&	0.08	\\
4	&	6.09898*	&	0.00	&	16.44	&	19.54	&	7.14	&	0.00	&	958.56	&	13.87	&	4.15	&	1.11	&	26.43	&	104.58	&	0.06	\\
5	&	5.30536*	&	0.00	&	21.77	&	0.41	&	10.84	&	0.50	&	1556.91	&	0.58	&	5.28	&	1.34	&	26.32	&	107.18	&	0.08	\\
10	&	3.37658*	&	0.00	&	81.66	&	0.04	&	39.94	&	0.00	&	4920.78	&	4.30	&	13.40	&	2.48	&	26.41	&	170.44	&	0.08	\\
15	&	2.86364*	&	0.00	&	264.69	&	2.49	&	92.27	&	1.58	&	8541.13	&	5.50	&	22.87	&	0.54	&	26.39	&	198.61	&	0.09	\\
20	&	2.57320	&	0.74	&	416.70	&	3.11	&	167.95	&	0.77	&	12,588.48	&	4.19	&	35.20	&	0.75	&	26.45	&	163.11	&	0.12	\\
25	&	2.38540	&	1.54	&	630.36	&	0.75	&	242.45	&	0.26	&	17,222.64	&	2.68	&	48.77	&	1.32	&	26.42	&	131.92	&	0.14	\\

\midrule
$E_\text{aver}$	&		&	0.51	&		&	3.29	&		&	0.39	&		&	3.91	&		&	1.22	&		&	117.80	&		\\
$t_\text{init}$	&		&		&	14.72	&		&	14.19	&		&	14.23	&		&	14.15	&		&	2.70	&		&	2.66	\\
$t_\text{total}$	&		&		&	645.08	&		&	256.64	&		&	17,236.88	&		&	62.92	&		&	213.71	&		&	3.44	\\

\bottomrule
\end{tabular}
\label{taulukko_protein}
} 
\end{table}

\begin{table}[ht!]
\centering
\caption{Summary of the results with Range Queries Aggregates ($\times 10^{14}$). Dimensions: $m$ = 200,000, $n$ = 7.} \vspace{0.25cm}
\resizebox{\textwidth}{!}{ 
\begin{tabular}{
    c c 
    c c
    c c
    c c
    c c
    c c
    c c
}
\toprule
$k$ & $f_\text{best}$ 
& \multicolumn{2}{c}{\textsc{Clust-Splitter}} 
& \multicolumn{2}{c}{\textsc{LMBM-Clust}} 
& \multicolumn{2}{c}{\textsc{DC-Clust}}
& \multicolumn{2}{c}{\textsc{Big-Clust}}
& \multicolumn{2}{c}{\textsc{Big-Means}}
& \multicolumn{2}{c}{\textsc{MiniBatchKMeans}} \\
\cmidrule(lr){3-4} \cmidrule(lr){5-6} \cmidrule(lr){7-8} \cmidrule(lr){9-10} \cmidrule(lr){11-12} \cmidrule(lr){13-14}
& & \multicolumn{1}{c}{$E_k$} & \multicolumn{1}{c}{$t_k$} 
& \multicolumn{1}{c}{$E_k$} & \multicolumn{1}{c}{$t_k$} 
& \multicolumn{1}{c}{$E_k$} & \multicolumn{1}{c}{$t_k$} 
& \multicolumn{1}{c}{$E_k$} & \multicolumn{1}{c}{$t_k$} 
& \multicolumn{1}{c}{$E_k$} & \multicolumn{1}{c}{$t_k$} 
& \multicolumn{1}{c}{$E_k$} & \multicolumn{1}{c}{$t_k$} \\
\midrule
2	&	16.39968	&	0.00	&	5.55	&	0.00	&	2.11	&	0.00	&	90.66	&	0.01	&	1.21	&	0.01	&	4.26	&	0.20	&	0.15	\\
3	&	8.22970	&	0.00	&	8.53	&	0.00	&	5.25	&	0.00	&	222.08	&	0.02	&	3.24	&	0.03	&	4.25	&	1.35	&	0.16	\\
4	&	5.06319*	&	0.00	&	12.11	&	0.00	&	7.44	&	0.00	&	339.45	&	0.03	&	4.93	&	0.03	&	4.27	&	3.32	&	0.12	\\
5	&	3.49938*	&	0.00	&	16.28	&	0.00	&	9.94	&	0.00	&	457.86	&	0.05	&	6.22	&	0.04	&	4.34	&	7.42	&	0.09	\\
10	&	1.35277*	&	0.00	&	47.23	&	0.01	&	25.34	&	0.01	&	1048.14	&	0.20	&	11.04	&	1.59	&	4.43	&	3.77	&	0.17	\\
15	&	0.93917	&	0.00	&	77.52	&	0.00	&	43.62	&	0.00	&	1616.27	&	1.17	&	15.18	&	0.76	&	4.39	&	3.36	&	0.21	\\
20	&	0.74585	&	0.14	&	109.98	&	0.17	&	59.34	&	0.21	&	2234.92	&	0.00	&	20.20	&	0.21	&	4.27	&	3.56	&	0.17	\\
25	&	0.62330*	&	0.00	&	158.28	&	1.40	&	73.62	&	0.46	&	2920.66	&	0.74	&	24.79	&	0.49	&	4.28	&	5.04	&	0.14	\\

\midrule
$E_\text{aver}$	&		&	0.02	&		&	0.20	&		&	0.09	&		&	0.28	&		&	0.40	&		&	3.50	&		\\
$t_\text{init}$	&		&		&	2.11	&		&	2.03	&		&	2.03	&		&	2.12	&		&	0.54	&		&	0.56	\\
$t_\text{total}$	&		&		&	160.39	&		&	75.66	&		&	2922.69	&		&	26.91	&		&	35.02	&		&	1.77	\\

\bottomrule
\end{tabular}
\label{taulukko_rangeque}
} 
\end{table}

\begin{table}[ht!]
\centering
\caption{Summary of the results with Skin Segmentation ($\times 10^9$). Dimensions: $m$ = 245,057, $n$ = 3.} \vspace{0.25cm}
\resizebox{\textwidth}{!}{ 
\begin{tabular}{
    c c 
    c c
    c c
    c c
    c c
    c c
    c c
}
\toprule
$k$ & $f_\text{best}$ 
& \multicolumn{2}{c}{\textsc{Clust-Splitter}} 
& \multicolumn{2}{c}{\textsc{LMBM-Clust}} 
& \multicolumn{2}{c}{\textsc{DC-Clust}}
& \multicolumn{2}{c}{\textsc{Big-Clust}}
& \multicolumn{2}{c}{\textsc{Big-Means}}
& \multicolumn{2}{c}{\textsc{MiniBatchKMeans}} \\
\cmidrule(lr){3-4} \cmidrule(lr){5-6} \cmidrule(lr){7-8} \cmidrule(lr){9-10} \cmidrule(lr){11-12} \cmidrule(lr){13-14}
& & \multicolumn{1}{c}{$E_k$} & \multicolumn{1}{c}{$t_k$} 
& \multicolumn{1}{c}{$E_k$} & \multicolumn{1}{c}{$t_k$} 
& \multicolumn{1}{c}{$E_k$} & \multicolumn{1}{c}{$t_k$} 
& \multicolumn{1}{c}{$E_k$} & \multicolumn{1}{c}{$t_k$} 
& \multicolumn{1}{c}{$E_k$} & \multicolumn{1}{c}{$t_k$} 
& \multicolumn{1}{c}{$E_k$} & \multicolumn{1}{c}{$t_k$} \\
\midrule
2	&	1.32236	&	0.00	&	0.45	&	0.00	&	0.59	&	0.00	&	69.61	&	0.02	&	1.12	&	0.00	&	3.28	&	0.08	&	0.16	\\
3	&	0.89362	&	0.00	&	0.88	&	0.00	&	1.34	&	0.00	&	130.41	&	0.03	&	1.70	&	0.01	&	3.36	&	5.20	&	0.13	\\
4	&	0.63998	&	8.59	&	2.61	&	8.59	&	2.22	&	0.00	&	183.75	&	0.04	&	2.53	&	1.92	&	3.26	&	3.90	&	0.20	\\
5	&	0.50203	&	0.00	&	3.14	&	0.00	&	3.08	&	0.00	&	243.22	&	1.36	&	3.16	&	1.60	&	3.31	&	10.48	&	0.11	\\
10	&	0.25121	&	4.99	&	6.30	&	13.37	&	6.87	&	0.00	&	505.56	&	7.88	&	7.44	&	5.58	&	3.47	&	9.16	&	0.14	\\
15	&	0.16688*	&	0.00	&	11.95	&	26.76	&	11.78	&	1.84	&	762.83	&	4.93	&	11.02	&	5.35	&	3.26	&	11.20	&	0.12	\\
20	&	0.12615	&	0.25	&	19.25	&	11.56	&	17.34	&	0.37	&	1030.64	&	5.76	&	15.04	&	3.95	&	3.27	&	10.70	&	0.12	\\
25	&	0.10228	&	3.64	&	28.03	&	14.43	&	23.31	&	0.00	&	1341.83	&	8.97	&	20.57	&	4.78	&	3.30	&	7.64	&	0.23	\\

\midrule
$E_\text{aver}$	&		&	2.18	&		&	9.34	&		&	0.28	&		&	3.62	&		&	2.90	&		&	7.30	&		\\
$t_\text{init}$	&		&		&	1.02	&		&	0.91	&		&	0.91	&		&	1.01	&		&	0.29	&		&	0.30	\\
$t_\text{total}$	&		&		&	29.05	&		&	24.22	&		&	1342.73	&		&	21.57	&		&	26.78	&		&	1.51	\\

\bottomrule
\end{tabular}
\label{taulukko_skin}
} 
\end{table}

\begin{table}[ht!]
\centering
\caption{Summary of the results with 3D Road Network ($\times 10^6$). Dimensions: $m$ = 434,874, $n$ = 3.} \vspace{0.25cm}
\resizebox{\textwidth}{!}{ 
\begin{tabular}{
    c c 
    c c
    c c
    c c
    c c
    c c
    c c
}
\toprule
$k$ & $f_\text{best}$ 
& \multicolumn{2}{c}{\textsc{Clust-Splitter}} 
& \multicolumn{2}{c}{\textsc{LMBM-Clust}} 
& \multicolumn{2}{c}{\textsc{DC-Clust}}
& \multicolumn{2}{c}{\textsc{Big-Clust}}
& \multicolumn{2}{c}{\textsc{Big-Means}}
& \multicolumn{2}{c}{\textsc{MiniBatchKMeans}} \\
\cmidrule(lr){3-4} \cmidrule(lr){5-6} \cmidrule(lr){7-8} \cmidrule(lr){9-10} \cmidrule(lr){11-12} \cmidrule(lr){13-14}
& & \multicolumn{1}{c}{$E_k$} & \multicolumn{1}{c}{$t_k$} 
& \multicolumn{1}{c}{$E_k$} & \multicolumn{1}{c}{$t_k$} 
& \multicolumn{1}{c}{$E_k$} & \multicolumn{1}{c}{$t_k$} 
& \multicolumn{1}{c}{$E_k$} & \multicolumn{1}{c}{$t_k$} 
& \multicolumn{1}{c}{$E_k$} & \multicolumn{1}{c}{$t_k$} 
& \multicolumn{1}{c}{$E_k$} & \multicolumn{1}{c}{$t_k$} \\
\midrule
2	&	49.13298	&	0.00	&	0.72	&	0.00	&	19.44	&	0.00	&	195.39	&	0.02	&	3.76	&	0.01	&	4.24	&	0.62	&	0.20	\\
3	&	22.77818	&	0.00	&	1.52	&	0.00	&	21.47	&	0.00	&	480.98	&	0.03	&	6.77	&	0.03	&	4.42	&	1.19	&	0.30	\\
4	&	13.52389	&	0.00	&	2.08	&	0.00	&	23.45	&	0.00	&	734.70	&	0.03	&	8.97	&	0.05	&	4.26	&	2.87	&	0.15	\\
5	&	8.82574	&	0.00	&	2.73	&	0.00	&	25.59	&	0.00	&	1001.58	&	0.04	&	11.14	&	0.08	&	4.27	&	3.81	&	0.25	\\
10	&	2.56661	&	0.00	&	10.00	&	0.00	&	35.64	&	0.02	&	2316.09	&	0.34	&	17.59	&	0.26	&	4.32	&	5.36	&	0.13	\\
15	&	1.27069	&	0.00	&	24.36	&	0.00	&	47.30	&	0.00	&	3686.33	&	0.79	&	22.94	&	0.60	&	4.27	&	8.37	&	0.25	\\
20	&	0.80865*	&	0.00	&	59.50	&	0.00	&	63.95	&	0.01	&	5077.50	&	0.68	&	28.97	&	0.71	&	4.28	&	11.04	&	0.23	\\
25	&	0.59258	&	1.60	&	92.55	&	0.00	&	87.08	&	3.78	&	6568.56	&	1.40	&	36.30	&	1.10	&	4.36	&	10.91	&	0.06	\\

\midrule
$E_\text{aver}$	&		&	0.20	&		&	0.00	&		&	0.48	&		&	0.42	&		&	0.35	&		&	5.52	&		\\
$t_\text{init}$	&		&		&	2.36	&		&	2.13	&		&	2.14	&		&	2.24	&		&	0.66	&		&	0.66	\\
$t_\text{total}$	&		&		&	94.91	&		&	89.20	&		&	6570.70	&		&	38.55	&		&	35.07	&		&	2.23	\\

\bottomrule
\end{tabular}
\label{taulukko_3droad}
} 
\end{table}

\begin{figure}[ht!]
    \centering
    \includegraphics[width=\textwidth]{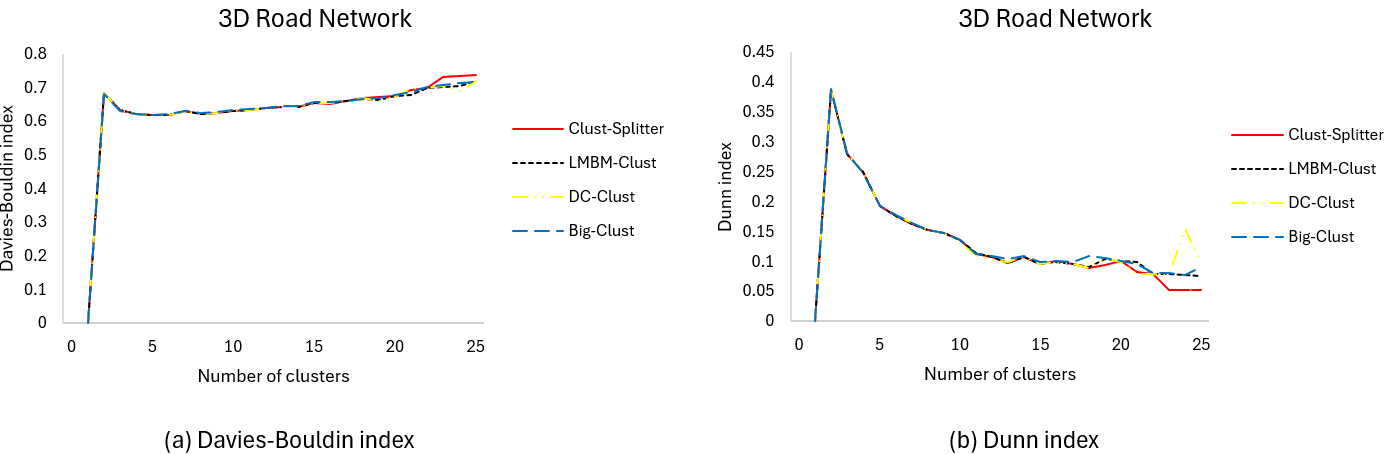}
    \caption{3D Road Network: Davies-Bouldin and Dunn validity indices vs. number of clusters.}
    \label{index_3droad}
\end{figure}

\begin{table}[ht!]
\centering
\caption{Summary of the results with Covertype ($\times 10^{11}$). Dimensions: $m$ = 581,012, $n$ = 10.} \vspace{0.25cm}
\resizebox{\textwidth}{!}{ 
\begin{tabular}{
    c c 
    c c
    c c
    c c
    c c
    c c
    c c
}
\toprule
$k$ & $f_\text{best}$ 
& \multicolumn{2}{c}{\textsc{Clust-Splitter}} 
& \multicolumn{2}{c}{\textsc{LMBM-Clust}} 
& \multicolumn{2}{c}{\textsc{DC-Clust}}
& \multicolumn{2}{c}{\textsc{Big-Clust}}
& \multicolumn{2}{c}{\textsc{Big-Means}}
& \multicolumn{2}{c}{\textsc{MiniBatchKMeans}} \\
\cmidrule(lr){3-4} \cmidrule(lr){5-6} \cmidrule(lr){7-8} \cmidrule(lr){9-10} \cmidrule(lr){11-12} \cmidrule(lr){13-14}
& & \multicolumn{1}{c}{$E_k$} & \multicolumn{1}{c}{$t_k$} 
& \multicolumn{1}{c}{$E_k$} & \multicolumn{1}{c}{$t_k$} 
& \multicolumn{1}{c}{$E_k$} & \multicolumn{1}{c}{$t_k$} 
& \multicolumn{1}{c}{$E_k$} & \multicolumn{1}{c}{$t_k$} 
& \multicolumn{1}{c}{$E_k$} & \multicolumn{1}{c}{$t_k$} 
& \multicolumn{1}{c}{$E_k$} & \multicolumn{1}{c}{$t_k$} \\
\midrule
2	&	13.41885	&	0.00	&	1.38	&	0.00	&	29.47	&	0.00	&	715.53	&	0.01	&	3.88	&	0.00	&	4.31	&	0.34	&	0.38	\\
3	&	9.52870	&	0.00	&	5.31	&	0.00	&	31.84	&	0.00	&	1535.64	&	0.02	&	6.99	&	0.01	&	4.28	&	5.15	&	0.48	\\
4	&	7.39484	&	0.07	&	13.22	&	0.07	&	34.78	&	0.07	&	2469.52	&	0.00	&	10.74	&	0.06	&	4.29	&	0.58	&	0.49	\\
5	&	5.89769	&	0.00	&	17.44	&	0.00	&	38.63	&	0.00	&	3315.48	&	0.03	&	13.52	&	0.02	&	4.28	&	3.13	&	0.57	\\
10	&	3.38780	&	2.13	&	55.33	&	0.00	&	60.42	&	0.00	&	8010.27	&	0.31	&	22.60	&	0.43	&	4.35	&	2.80	&	0.72	\\
15	&	2.46689*	&	0.00	&	106.64	&	0.00	&	81.69	&	0.87	&	12,961.38	&	0.26	&	31.35	&	0.46	&	4.28	&	4.49	&	0.57	\\
20	&	2.03496	&	0.39	&	178.25	&	0.05	&	114.14	&	0.00	&	18,125.08	&	0.29	&	42.45	&	0.60	&	4.30	&	2.17	&	0.70	\\
25	&	1.73617	&	0.44	&	274.78	&	0.00	&	154.13	&	0.03	&	23,422.95	&	0.37	&	54.60	&	0.59	&	4.31	&	3.75	&	0.70	\\

\midrule
$E_\text{aver}$	&		&	0.38	&		&	0.01	&		&	0.12	&		&	0.16	&		&	0.27	&		&	2.80	&		\\
$t_\text{init}$	&		&		&	6.42	&		&	6.22	&		&	6.22	&		&	6.26	&		&	1.87	&		&	1.97	\\
$t_\text{total}$	&		&		&	281.20	&		&	160.34	&		&	23,429.17	&		&	60.86	&		&	36.26	&		&	6.58	\\

\bottomrule
\end{tabular}
\label{taulukko_covertype}
} 
\end{table}

\end{document}